\documentclass{article}

\usepackage{subcaption}

\usepackage{microtype}
\usepackage{graphicx}
\usepackage{booktabs} % for professional tables
\usepackage{hyperref}       % hyperlinks
\usepackage{url}            % simple URL typesetting
\usepackage{amsfonts}       % blackboard math symbols
\usepackage{nicefrac}       % compact symbols for 1/2, etc.
\usepackage{microtype}      % microtypography
\usepackage{xcolor}         % colors
\usepackage{amsmath}
\usepackage{mathtools}
\usepackage{hyperref}
\usepackage{adjustbox}
\usepackage{graphicx} % For including images
\usepackage{algorithm, algpseudocode}
\usepackage{hyperref}
\usepackage{hyperref}

\usepackage[accepted]{icml2025}

\usepackage{amsmath}
\usepackage{amssymb}
\usepackage{mathtools}
\usepackage{amsthm}
\usepackage{adjustbox}

\usepackage[capitalize,noabbrev]{cleveref}

\theoremstyle{plain}

\theoremstyle{definition}

\theoremstyle{remark}

\theoremstyle{remark} % Or 'plain' or 'definition' depending on your goal
\newtheorem*{hypothesis*}{\textsc{Hypothesis}}

\newtheoremstyle{italic-hypothesis}
  {\topsep}   % space above
  {\topsep}   % space below
  {\itshape}  % body font
  {}          % indent
  {\scshape}  % head font (small caps)
  {.}         % punctuation after head
  { }         % space after head
  {\thmname{#1}\thmnumber{ #2}\thmnote{ (#3)}} % head spec

\theoremstyle{italic-hypothesis}

\newcommand{\x}{\bold{x}}
\newcommand{\y}{\bold{y}}
\DeclareMathOperator*{\argmax}{arg\,max}

\usepackage[textsize=tiny]{todonotes}

\icmltitlerunning{Down with the Hierarchy: The `H' in HNSW Stands for ``Hubs"}

\begin{document}

\twocolumn[
\icmltitle{Down with the Hierarchy: The `H' in HNSW Stands for ``Hubs"}




\begin{icmlauthorlist}
\icmlauthor{Blaise Munyampirwa}{arg}
\icmlauthor{Vihan  Lakshman}{yyy}
\icmlauthor{Benjamin Coleman}{goog}
\end{icmlauthorlist}

\icmlaffiliation{arg}{Argmax Inc., Mountain View, CA}
\icmlaffiliation{yyy}{MIT CSAIL, Cambridge, MA}
\icmlaffiliation{goog}{Google DeepMind, Mountain View, CA}

\icmlcorrespondingauthor{Blaise Munyampirwa}{blaisemunyampirwa@gmail.com}

\icmlkeywords{Vector Databases, Approximate Near-Neighbor Search, HNSW}

\vskip 0.3in
]



\printAffiliationsAndNotice{}  % leave blank if no need to mention equal contribution

\begin{abstract}
Driven by recent breakthrough advances in neural representation learning, approximate near-neighbor (ANN) search over vector embeddings has emerged as
a critical computational workload. With the introduction of the seminal Hierarchical Navigable Small World (HNSW) algorithm, graph-based indexes have established themselves as the overwhelmingly dominant paradigm for efficient and scalable ANN search. As the name suggests, HNSW searches a layered hierarchical graph to quickly identify neighborhoods of similar points to a given query vector. But is this hierarchy even necessary? A rigorous experimental analysis to answer this question would provide valuable insights into the nature of algorithm design for ANN search and motivate directions for future work in this increasingly crucial domain. We conduct an extensive benchmarking study covering more large-scale datasets than prior investigations of this question. We ultimately find that a flat navigable small world graph graph retains all of the benefits of HNSW on high-dimensional datasets, with latency and recall performance essentially \emph{identical} to the original algorithm but with less memory overhead. Furthermore, we go a step further and study \emph{why} the hierarchy of HNSW provides no benefit in high dimensions, hypothesizing that navigable small world graphs contain a well-connected, frequently traversed ``highway" of hub nodes that maintain the same purported function as the hierarchical layers. We present compelling empirical evidence that the \emph{Hub Highway Hypothesis} holds for real datasets and investigate the mechanisms by which the highway forms. The implications of this hypothesis may also provide future research directions in developing enhancements to graph-based ANN search. 

\end{abstract}

\section{Introduction}
Near neighbor search is a fundamental problem in computational geometry that lies at the heart of countless practical applications. From industrial-scale recommendation  \cite{feng2022recommender} to retrieval-augmented generation \cite{Lewis2020RetrievalAugmentedGF} and even to computational biology \cite{zhao2024gsearch}, numerous data-intensive tasks utilize similarity search at some location in the stack.
As a result, similarity indexes are very well-studied
\cite{Guo2019AcceleratingLI, Malkov2016EfficientAR, Johnson2017BillionScaleSS, Aguerrebere2023Intel, Subramanya2019DiskANNF} with multiple large-scale benchmarks and leaderboards to compare techniques 
\cite{Aumller2018ANNBenchmarksAB, Simhadri2022}. 

Historically, the state-of-the-art for near neighbor search involved constructing sophisticated tree-based data structures, such as $kd$-trees \cite{bentley1975multidimensional} and cover trees \cite{beygelzimer2006cover}, that guaranteed exact solutions while avoiding a brute-force examination of all points. However, the recent advent of large-scale neural representation learning, including large language models (LLMs), places a significant strain on these classical methods that were developed to target a much lower-dimensional search space. In response, the community has turned to approximate search methods. While alternative approximate indexing methods such as locality-sensitive hashing \cite{indyk1998approximate} and product quantization \cite{jegou2010product}, have garnered significant interest, graph-based approaches generally achieve the strongest performance on established ANN benchmarks \cite{Aumller2018ANNBenchmarksAB, Simhadri2022}. Introduced in 2016, the Hierarchical Navigable Small World (HNSW) algorithm \cite{Malkov2016EfficientAR}, emerged as one of the first high-performance graph-based search indexes at scale and still enjoys immense popularity to this day with over 4300 Github stars\footnote{\url{https://github.com/nmslib/hnswlib}} and deployments in major commercial search systems such as Apache Lucene \cite{xian2024vector} and Pinterest \cite{Engineering_2021}. 

As the name implies, a core feature of the HNSW index is its hierarchically layered graph akin to a skip list \cite{pugh1990skip} where the search process iteratively traverses through graphs of increasing density before converging to a neighborhood of similar points in the final graph layer. By drawing intuition from skip lists, the HNSW authors argue that the initial coarse graph layers allow for efficiently identifying the neighborhood of similar points in the collection through fewer overall comparisons. 

Despite its popularity, HNSW has notable scalability issues; the hierarchical structure adds significant memory overhead and, as noted in \cite{Malkov2016EfficientAR}, can reduce throughput in distributed settings compared to flat NSW graphs. Although this overhead is often justified by the latency benefits of graph-based indexes, recent work questions whether the hierarchy is still necessary. \cite{lin2019graph} report that the hierarchy only improves performance for low-dimensional data ($d < 32$), and \cite{coleman2022graph} reach similar conclusions in their ablation study. These observations highlight the need for a more rigorous investigation into whether the hierarchy remains useful in modern, high-dimensional workloads.

Perhaps most importantly, we still have no satisfactory understanding of \emph{why hierarchy does not help}.
Hierarchical structures are a mainstay of algorithm design, where a common trick is to reduce an $O(n)$ search process to a sublinear one by traversing a (balanced) hierarchy \cite{pugh1990skip, guibas1978dichromatic, mikolov2013distributed, cormen2022introduction}. Arguably, it is counterintuitive for this idea to fail to hold in the context of high-dimensional similarity search -- especially when we have strong positive results that hierarchy \emph{helps} in low dimensions \cite{beygelzimer2006cover, dolatshah2015ball, ram2019revisiting, lin2019graph}. Thus, an exhaustive benchmark and deeper analysis into the necessity of the hierarchy in HNSW would shed further light on the nature of algorithm design in high-dimensional spaces and thus may be of independent interest to the community as well. 

\begin{figure}[t]
\begin{center}
\centerline{\includegraphics[width=2.2in]{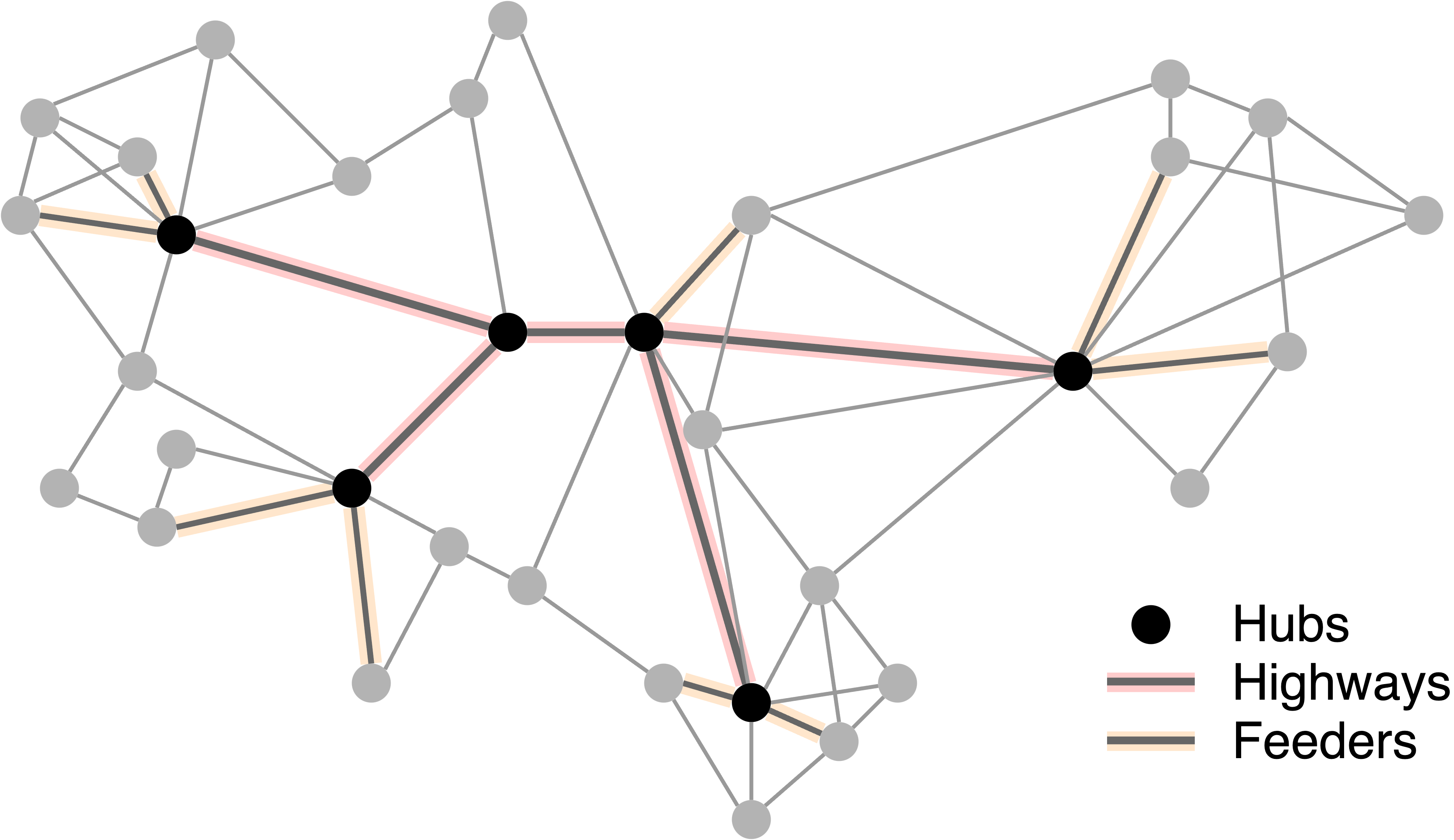}}
\end{center}
\caption{We hypothesize that, in high dimensions, graph-based ANN indexes naturally form a ``highway-feeder'' structure, where a small subset of nodes and edges are easily reached, well-connected, and heavily traversed.}
\label{fig:hub_highway_feeder}
\end{figure}

\subsection{Contributions}

In this paper, we study whether the hierarchical component of HNSW is truly necessary. Our central research question is, \textit{``Can we achieve the same performance on large-scale benchmarks with simply a flat navigable small world graph?''} To that end, we organize the paper into two parts:

\textbf{Benchmarking the hierarchy:} We rigorously benchmark HNSW to understand whether the hierarchy is necessary. To do so, we reproduce and extend the hierarchy ablations of previous studies, finding that, on high-dimensional vector datasets, it is indeed beneficial to remove the `H' from HNSW. 

\textbf{Why does hierarchy not help?} We hypothesize that the hierarchy benefits decrease in high-dimensions due to \textit{hubness}.
Hubness is a high-dimensional phenomenon that causes a skewed distribution in the near-neighbor lists of search queries~\cite{Radovanovi2010HubsIS}.
We hypothesize that hubness leads to preferential attachment in the similarity search graph, inducing the formation of \emph{easily-traversed highways} that connect disparate regions of the graph.
This hypothesis, which we call the \textit{Hub Highway Hypothesis}, explains why we no longer need the hierarchy in high dimensions; we can simply traverse the intrinsic highway structure that naturally forms in high-dimensional spaces. Our results ultimately show that hubness is responsible for driving the connectivity of similarity search graphs.
This insight opens up exciting new research directions in graph construction, link pruning, and graph traversal.

% \footnote{Specifically, that the benefits of sophisticated initialization methods will decay under hub-style preferential attachment.}

Our specific contributions are as follows.
\begin{itemize}
    \item We release an implementation for a flattened version of HNSW, called FlatNav\footnote{\url{https://github.com/BlaiseMuhirwa/flatnav}}, that reaches performance parity with the original hierarchical version with considerable memory savings. To our knowledge, \texttt{flatnav} is the only actively maintained, high-performance NSW search library in the open-source ecosystem and thus fills a crucial void in the similarity search community.  
    \item We demonstrate that hierarchy does not improve performance in either the median or tail latency case by building HNSW and FlatNav indexes for 13 popular benchmark datasets ranging in size from 1 million to 100 million vectors.
    % \item We conduct an analysis of hubness phenomena in high-dimensional metric spaces and the resulting HNSW graphs, finding strong empirical support for the hub-highway hypothesis. 
    \item We present strong scientific evidence for the hub-highway hypothesis, drawing empirical support from analysis of hubness phenomena in high-dimensional metric spaces and the resulting HNSW graphs. 
\end{itemize}

\textbf{Practical implications:} Our benchmarks reveal that HNSW can be significantly optimized for modern high-dimensional embedding workloads. For instance, as we show in Table~\ref{tab:datasets-memory} in the appendix, our implementation saves roughly $38\%$ and $39\%$ of peak memory consumption during index construction on two Big-ANN benchmark datasets compared to \texttt{hnswlib} (and sizable further headroom is likely). Our results confirm the folklore of the similarity search community, conclusively demonstrating that we can remove the hierarchy on high-dimensional inputs with impunity.

\section{Related Work}

\subsection{Near-Neighbor Benchmarks}

Recent years have seen the emergence of large-scale benchmarks for the $k$-NNS problem. ANN Benchmarks \cite{Aumller2018ANNBenchmarksAB} established the first standard evaluation framework, expanding over time to cover 30+ methods across 9 datasets. However, the ANN Benchmark datasets are relatively small, with around one million points. To better reflect real-world scale, Big ANN Benchmarks \cite{Simhadri2022} introduced five billion-scale datasets. Nevertheless, these benchmarks emphasize overall throughput and omit tail latency metrics like the 99\textsuperscript{th} percentile, which are essential for evaluating hierarchical components such as those in HNSW.

\textbf{Hierarchy Studies:} Many top-performing graph-based ANN algorithms, including HNSW \cite{Malkov2016EfficientAR}, ONNG \cite{iwasaki2018optimization}, PANNG \cite{iwasaki2016pruned}, and HCNNG \cite{munoz2019hierarchical}, rely on hierarchical structures. However, recent work has questioned this design. \cite{dobson2023scaling} show that HNSW can underperform both HCNNG (with a shallower hierarchy) and DiskANN \cite{Subramanya2019DiskANNF} (which lacks one). \cite{lin2019graph} find hierarchy helpful only in low-dimensional synthetic settings ($d < 32$), but study few real-world datasets and do not explain the observed failure modes. We aim to address these gaps by reproducing prior results \cite{Malkov2016EfficientAR, lin2019graph} with our own implementation and extending the analysis to larger datasets with a focus on when and why hierarchy matters.

\subsection{Hubness in High Dimensional Spaces}

Astute readers might observe that the HNSW graph construction algorithm does not explicitly enforce the small world property and instead adds edges between nodes based on their proximity in the metric space. The connection between proximity and the small world property arises due to \emph{hubness}.

Hubness is a property of high-dimensional metric spaces where a small subset of points (the ``hubs'') occur a disproportionate number of times in the near-neighbor lists of other points in the dataset \citep{Radovanovi2010HubsIS}. In other words, a small fraction of nodes are highly connected to other points in the near-neighbor graph. The concentration of distance and measure in high-dimensional spaces provide good intuition for how hubness can arise in a datasets. The concentration of distances is a well-studied phenomenon where the expected $\ell_2$ distance between independent and identically distributed (i.i.d) vectors grows with $\sqrt{d}$ while the variance tends to a constant as $d$ approaches infinity \citep{Talagrand1994ConcentrationOM}. As a result, the $\ell_2$ distance loses its discriminative power as $d$ increases, a fact which also holds for $\ell_p$ and fractional norms \citep{Franois2007TheCO}. Concentration of measure is a high-dimensional property where random distributions have most of their mass near the boundary of their domain.
Taken together, these facts suggest that hubs will form at extrema of high-dimensional datasets - a result which holds true empirically \citep{Low2013TheHP}.

Due to undesirable consequences of the hubness phenomenon, such as poor clustering quality, a large body of work has focused on hubness reduction strategies. For instance, \citep{ZelnikManor2004SelfTuningSC} introduced local scaling which scales distances $d(\x,\y)$ by accounting for local neighborhood information. Interestingly, our work stands in contrast to this literature on hubness reduction by presenting a case study where hubs provide tangible value in an algorithmic setting, namely in accelerating greedy traversal in near neighbor proximity graphs. This result may be of independent interest to machine learning and algorithms researchers as well.

\section{FlatNav Benchmarking Experiments} \label{latency-benchmarks}

In this section, we report the results of our benchmarking study comparing the performance of flat HNSW search to hierarchical search on a suite of standard high-dimensional benchmark datasets drawn from real machine learning models. We fix the implementation in our experimental design such that the \emph{same code} is used to construct the indexes. In particular, we use the \texttt{hnswlib} library as our baseline HNSW implementation. To benchmark the flat NSW index performance, we extract the bottom layer from \texttt{hnswlib} after constructing the full hierarchical graph and reimplement HNSW's greedy search heuristic over the flat graph via \texttt{flatnav}. 

One natural question that arises from this experimental design is whether the hierarchical component of HNSW might still be useful during \emph{construction} even if the base layer suffices for \emph{search}. We find that there is no difference in performance between these settings. Specifically, we include additional results comparing HNSW to flat graphs constructed from scratch (without the hierarchy) in Appendix \ref{extended-bench} where we see identical results to those reported in this section of the paper. In this section, we focus on extracting the base layer from the \texttt{hnswlib} graph to reduce any potential confounding effects from implementation differences. Nevertheless, we obtain identical results regardless of how the base graph is constructed. 

\subsection{Datasets and Compute} \label{sec:main-datasets}

We utilize the benchmark datasets released through the popular leaderboards ANN Benchmarks \cite{Aumller2018ANNBenchmarksAB} (MIT Licensed) and Big ANN Benchmarks (MIT Licensed) \cite{Simhadri2022}. The specific datasets and their associated statistics are presented in Table \ref{tab:datasets-exps}. For the Big ANN Benchmark datasets, we consider both the 10M and 100M collection of vectors, for which the ground truth near neighbors have previously been computed and released. We did not experiment with the largest Big ANN datasets with 1 billion vectors since constructing HNSW indexes at this scale requires over 1.5TB of RAM, which exceeded our compute resources. In this section, we include our benchmarking results for the four 100M-scale datasets available through Big ANN Benchmarks. We see that our flat HNSW implementation achieves performance parity with the hierarchical HNSW implementation. 

For our benchmarks on datasets consisting of fewer than 100M vectors in the collection, we use an AWS c6i.8xlarge instance with an Intel Ice Lake processor and 64GB of RAM. We selected this particular public cloud instance to facilitate accessible reproducibility of our experiments. For the 100M-sized large-scale experiments, we use a cloud server equipped with an AMD EPYC 9J14 96-Core Processor and 1 TB of RAM. 

\subsection{Latency Results}
\subsubsection{BigANN Benchmarks \cite{Simhadri2022}}
\begin{figure}[htbp]
    \centering
    \includegraphics[width=0.5\textwidth]{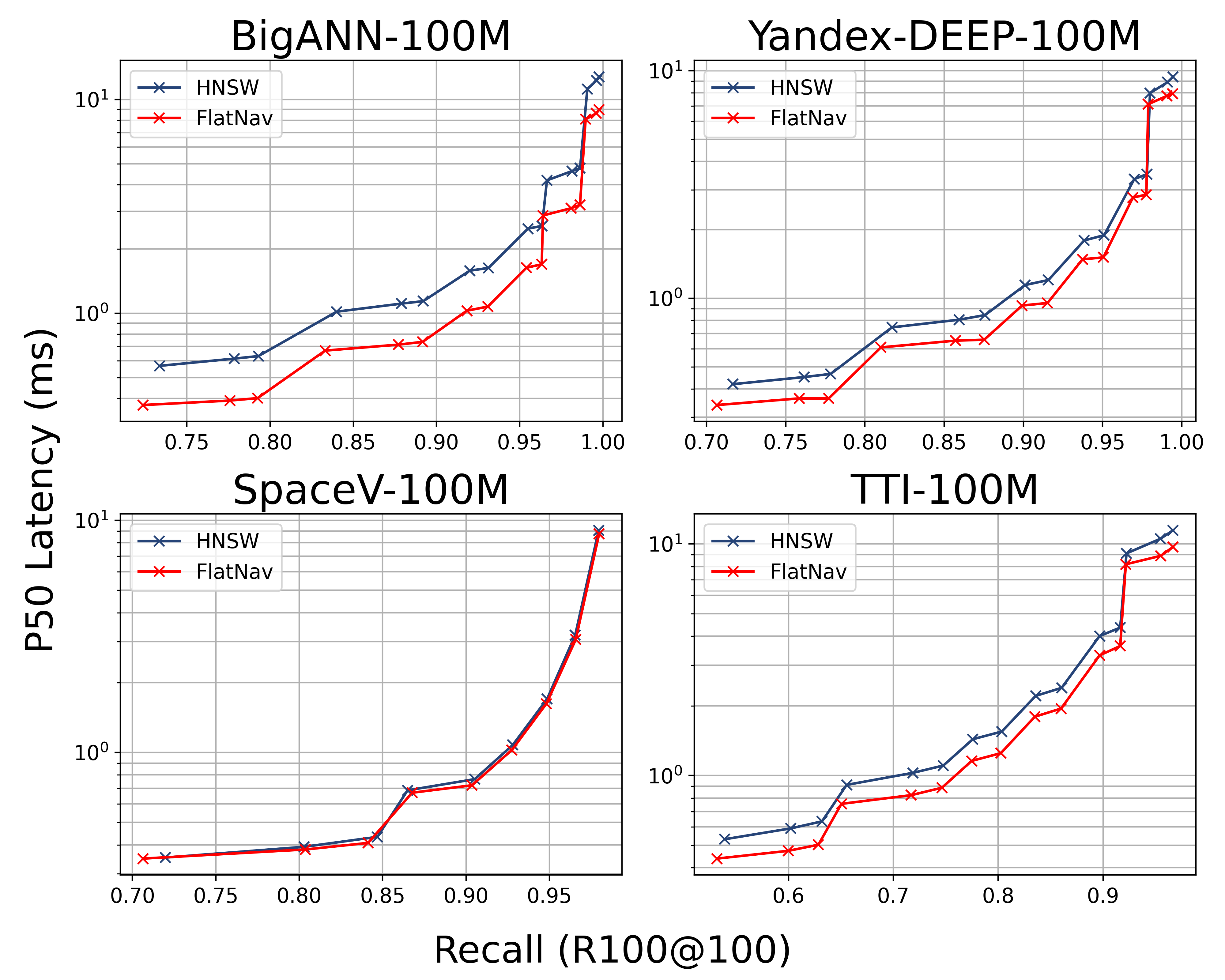}
    \caption{p50 Latency vs. Recall. FlatNav performs nearly identically to HNSW.}
    \label{fig:latency-p50}
    
    \includegraphics[width=0.5\textwidth]{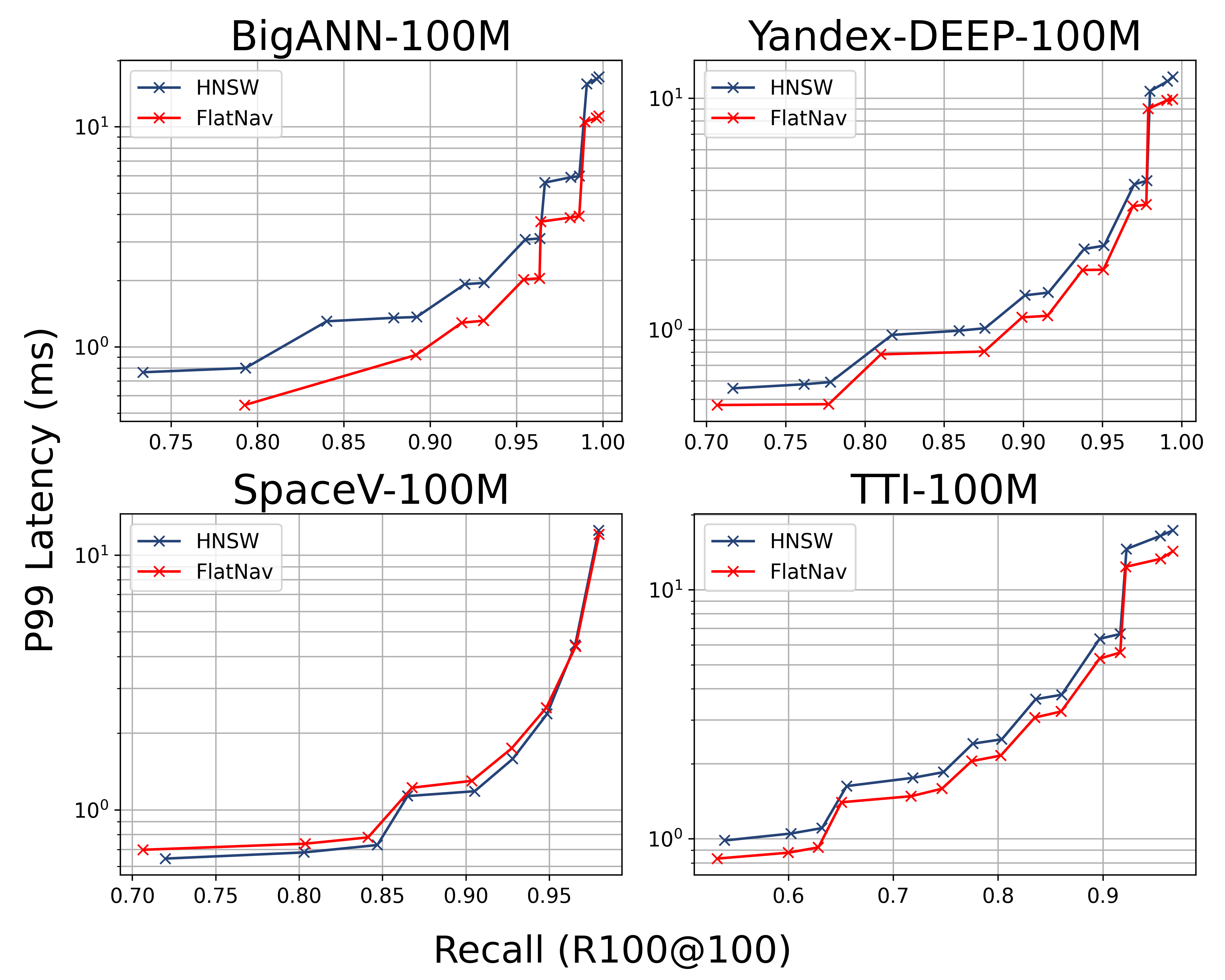}
    \caption{p99 Latency vs. Recall. FlatNav performs nearly identically to HNSW.}
    \label{fig:latency-p99}
    
    %\caption{Latency vs. Recall for HNSW and FlatNav over four 100M-scale benchmark datasets with dimensionality between 96 and 200. FlatNSW and HNSW show essentially identical performance in both the median and p99 cases.}
    \label{fig:latency-combined}
\end{figure}

In Figures \ref{fig:latency-p50} and \ref{fig:latency-p99}, we compare latency metrics for HNSW and FlatNav at the 50th and 99th percentile for the four 100M datasets from BigANN benchmarks listed in Table \ref{tab:datasets-exps}. All of our results support the conclusion that \texttt{flatnav} achieves nearly identical performance to \texttt{hnswlib}. 

From the results in Figures \ref{fig:latency-p50} and \ref{fig:latency-p99}, we observe that there is no consistent and discernable gap between FlatNav and HNSW in both the median and tail latency cases. These results suggest that the hierarchical structure of HNSW provides no tangible benefit on practical high-dimensional embedding datasets.  

% We further notice a similar pattern when we measure throughout and the number of distance computations, which we include in the Appendix. 

% For the SpaceV and Yandex Deep 100M datasets, we see that HNSW's p99.9 latency is roughly 1ms faster in the low-recall regime and this gap is not present in even the p99 case. However, this difference is not statistically significant and we do not see this behavior consistently in many of our other benchmark datasets (included in the appendix). 

\subsubsection{ANN Benchmarks \cite{Aumller2018ANNBenchmarksAB}}

\begin{figure}[htbp]
  \centering

  % Row 1: Two latency plots
    \centering
    \includegraphics[width=0.5\textwidth]{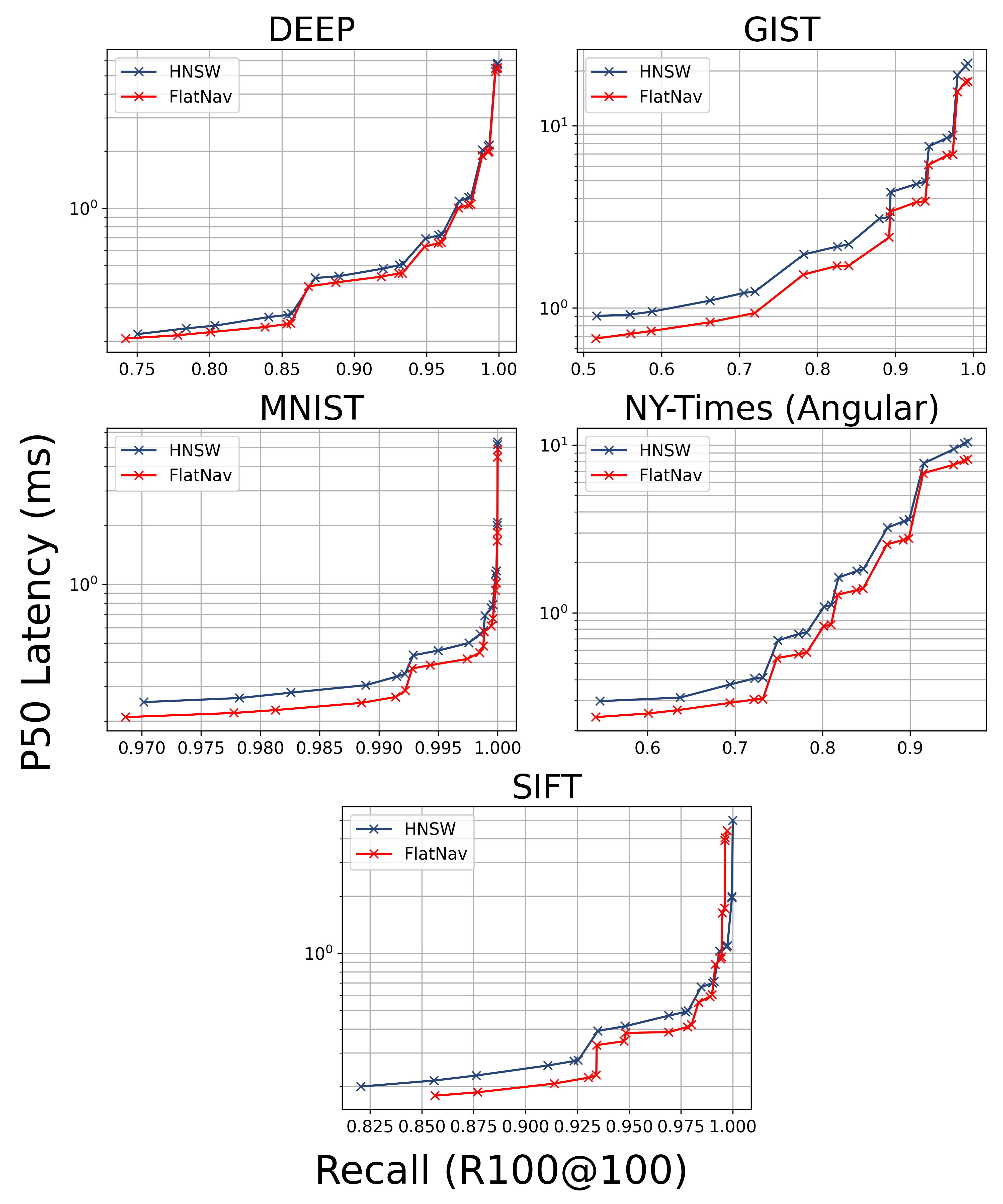}
    \caption{p50 Latency vs. Recall. FlatNav performs nearly identically to HNSW.}
    \label{fig:annbench-p50}
  \hfill
    \centering
    \includegraphics[width=0.5\textwidth]{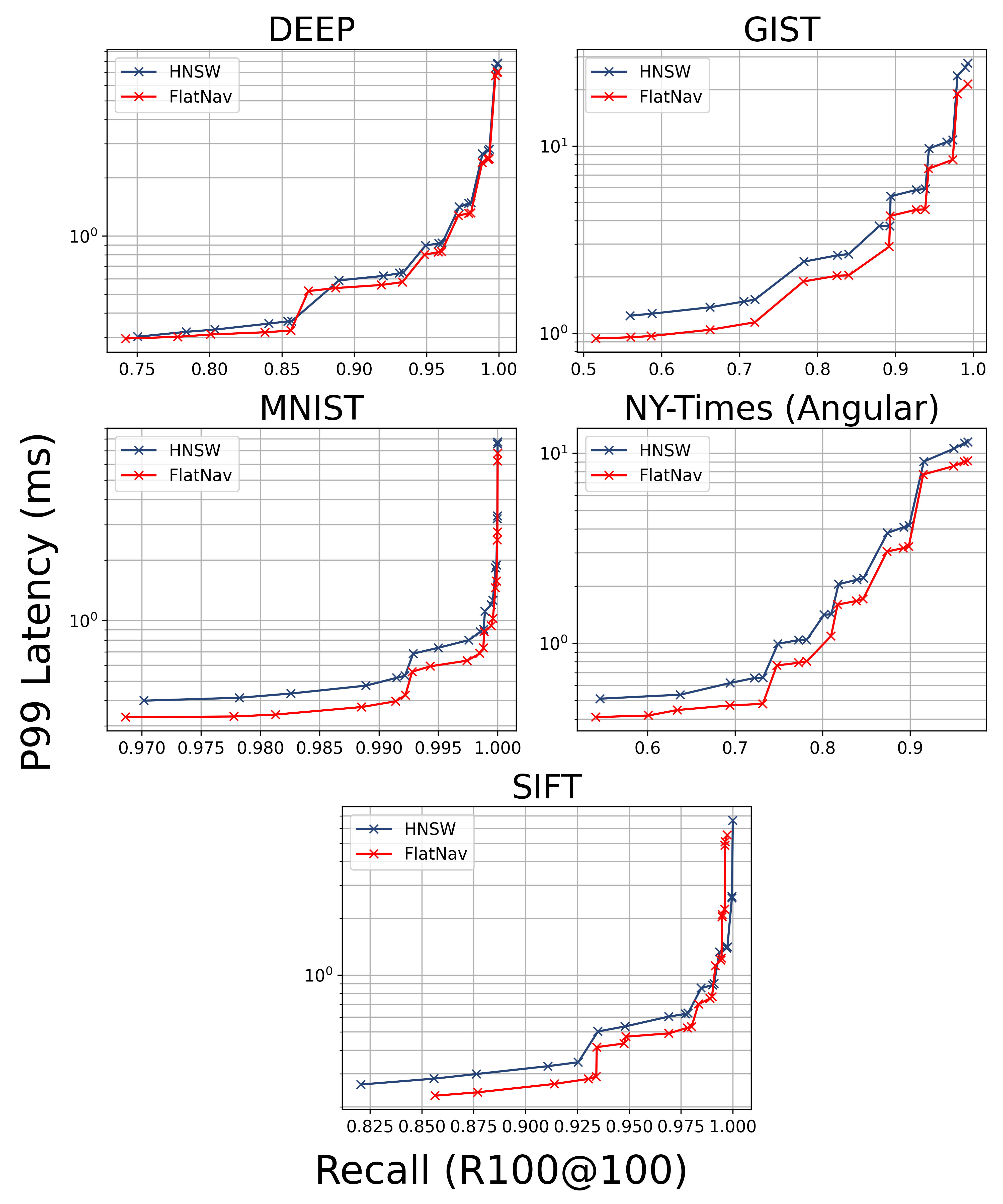}
    \caption{p99 Latency vs. Recall. FlatNav performs nearly identically to HNSW.}
    \label{fig:annbench-p99}

  \vspace{1em} % Small vertical gap between rows

\end{figure}

\begin{figure}[t]
%    \centering
    \includegraphics[width=0.5\textwidth]{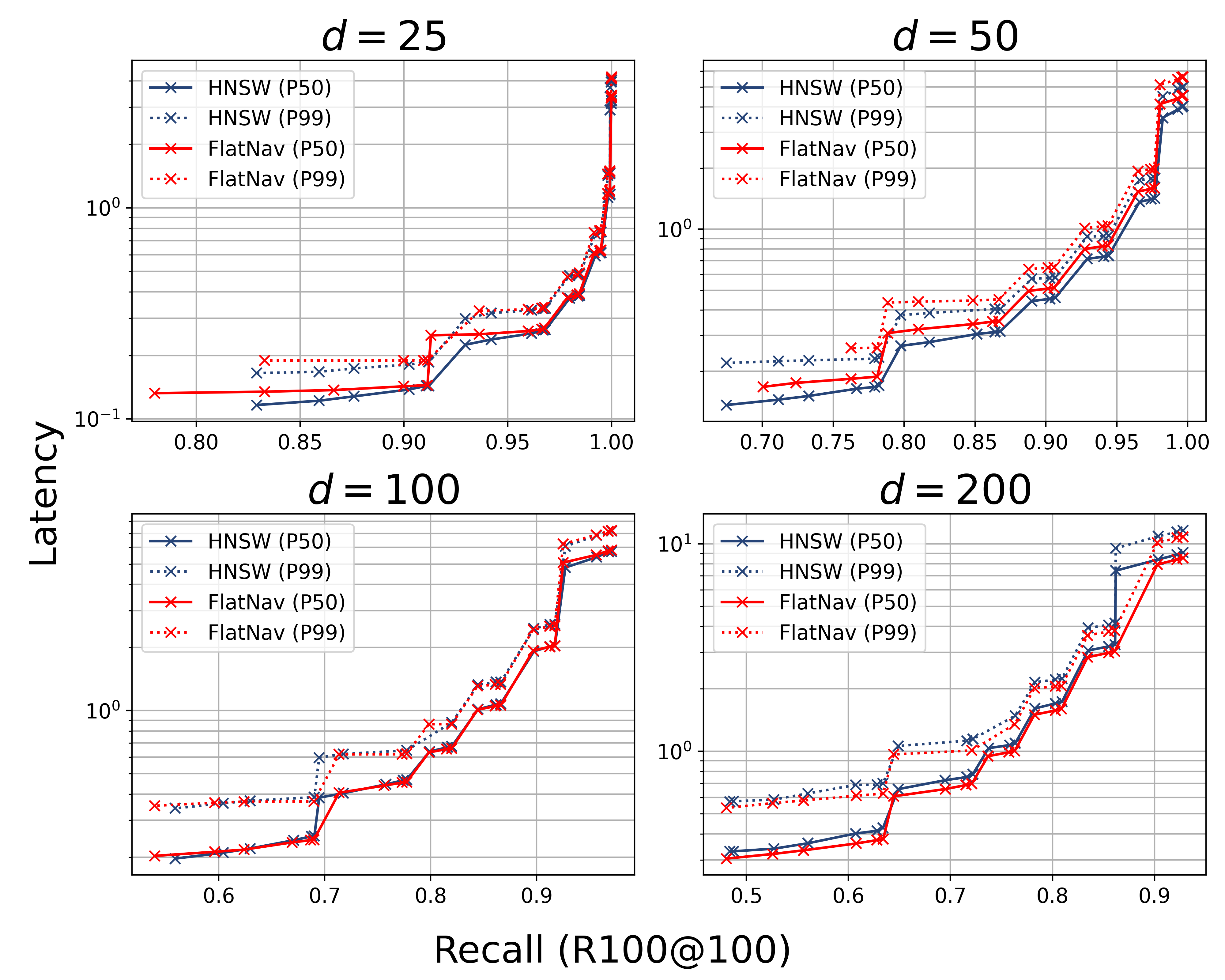}
    \caption{p50 and p99 Latency vs. Recall for HNSW and FlatNav over GloVe datasets.}
    \label{fig:glove-latency}
  \end{figure}

We repeat the same experimental setup comparing HNSW and FlatNav on the ANN Benchmark datasets listed in Table \ref{tab:datasets-exps}. In Figures \ref{fig:annbench-p50} and \ref{fig:annbench-p99}, we report the p50 and p99 latency of all of the non-GloVe ANN Benchmarks. Although these datasets are smaller in scale than the BigANN Benchmarks, we still see no discernible difference in latency between HNSW and FlatNav which supports our hypothesis that the vector dimensionality and not the size of the collection is the main driver of eliminating the need for hierarchical search in small world graphs. We see further evidence of this idea in Figure \ref{fig:glove-latency}, which confirms that there is no clear performance benefit provided by the hierarchy of HNSW. 

  % Row 2: GloVe latency + node access distribution

  \begin{figure}
%    \centering
\includegraphics[width=0.45\textwidth]{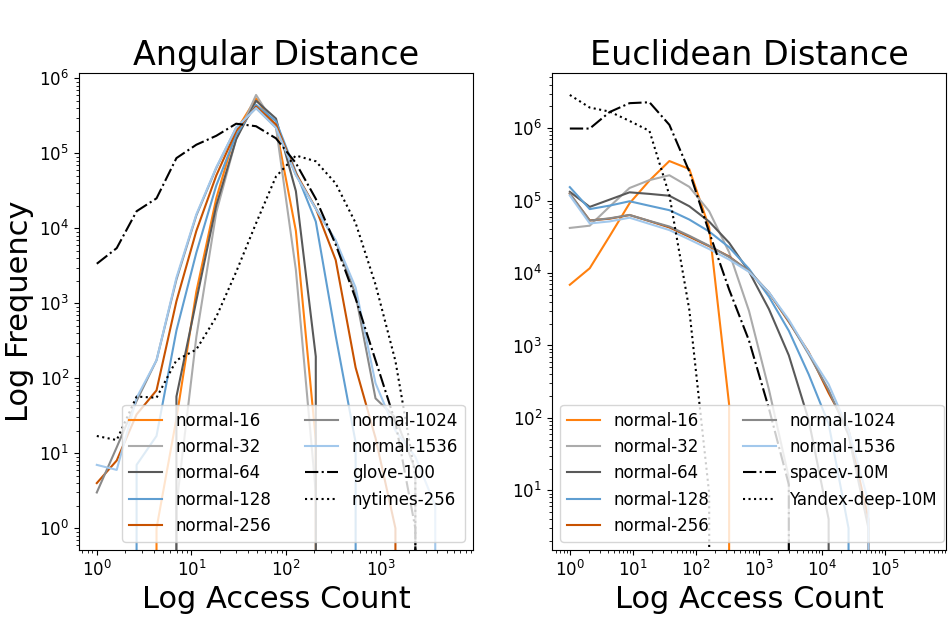}
    \caption{Log-normalized Node access count distribution $P_m(\x_i)$ for datasets using angular (left) and $l_2$ (right) distances.}
    \label{fig:combined-kde-plots}
  %\caption{Latency and access distribution comparisons for ANN search across datasets.}
  \label{fig:combined-latency-access}
\end{figure}

\section{The Hub Highway Hypothesis}
\label{section:hubs}

We now turn our attention to studying why the hierarchy appears to provide no benefit in the search process. In our experiments, we observed that a small fraction of nodes appear in the set of near neighbors for a disproportionate number of other vectors. We thus conjecture that the hub structure prevalent in high-dimensional data performs the same functional role as the hierarchy.

\begin{hypothesis*}[Hub Highways]
\itshape In high-dimensional metric spaces, $k$-NN proximity graphs form a highway routing structure where a small subset of nodes are well-connected and heavily traversed, particularly in the early stages of graph search.
\end{hypothesis*}

We remark that the existence of hub nodes in high-dimensional space is not a new observation \cite{Radovanovi2010HubsIS}. The novelty of our hypothesis lies in connecting the idea of hubness to the notion of accelerating near neighbor search in ANN proximity graphs. In particular, we conjecture that near neighbor queries over proximity graphs in high dimensions often spend the majority of their time visiting hub nodes early on in the search process before converging to a local neighborhood of near neighbors. This procedure succeeds because hub nodes are very well connected to other parts of the graph and thereby efficiently route queries to the appropriate neighborhood in much the same manner that the layered hierarchy purports to do. 

In the remainder of this section, we present an experimental design and a series of results that provide empirical evidence in the affirmative for the existence of such a highway routing mechanism amongst hubs in navigable small world graphs. 

\subsection{Methodology}

\textbf{Argument sketch:} We will demonstrate the \textit{Hub Highway Hypothesis} by providing empirical evidence for the following claims.
\begin{enumerate}
    \item Some nodes are visited by queries much more frequently than others. The relative popularity of these \textit{hub nodes} is explained by the hubness phenomenon that arises in high dimensions.
    \item The hub nodes form a well-connected subgraph of hubs (the \textit{highway} network).
    \item Queries visit many hub nodes early in the search process, before visiting less well-traversed neighborhoods.
\end{enumerate}

\textbf{Empirical measures of hubness:} To support the first part of our argument, we require a formal characterization of hubness. Following \cite{Radovanovi2010HubsIS}, let $\x, \x_1, \ldots, \x_n$ be vectors drawn from the same probability distribution supported on $\mathcal{S} \subseteq \mathbb{R}^d$, and let $\phi : \mathcal{S} \times \mathcal{S} \to \mathbb{R}$ be a distance function. For $1 \leq i, k \leq n$, define $p_{i,k}(\x) = 1$ if $\x$ is among the $k$-NN set of $\x_i$ under $\phi$, and $0$ otherwise. Define $N(\x) \coloneqq \sum_{i=1}^{n} p_{i,k}(\x)$, the number of vectors $\x_i$ for which $\x$ appears in their $k$-NN set.

Given a dataset $\mathcal{D}$, the values $\{N(\x)\}_{\x \in \mathcal{D}}$ form a distribution $N_k$, whose skewness is given by 
$S_{N_k} = \mathbb{E}[(N_k - \mu_{N_k})^3] \, / \, \sigma_{N_k}^3$.
We use $S_{N_k}$ to measure the hubness of a dataset.

% Following \citep{Radovanovi2010HubsIS}, let $\x, \x_1, \hdots, \x_n $ be a collection of vectors drawn from the same probability distribution with support $\mathcal{S} \subseteq \mathbb{R}^d$, and let $\phi: \mathcal{S} \times \mathcal{S} \to \mathbb{R}$ be a distance function.

% Furthermore, for $1 \leq i, k \leq n$, let $p_{i,k}$ be defined by
% \[
% p_{i,k}(\x) =
% \begin{cases} 
%       1 & \text{if $\x$ is among the $k$-NN set of $\x_i$ under $\phi$}  \\
%       0 & \text{otherwise } 
%    \end{cases}
% \]

% Now let $N(\x)$ be the random variable defined by 
% \[
% N(\x) \coloneqq \sum_{i=1}^{n} p_{i,k}(\x)
% \]
% which represents the number of vectors that have $\x$ included in their $k$-nearest neighbors.

% For any dataset $\mathcal{D}$, we can compute $N(\x), \x \in \mathcal{D}$, which yields a discrete distribution $N_k$. We are interested in the skewness of this distribution, given by 
% \[
% S_{N_{k}} = \frac{\mathbb{E}\left[\left(N_k - \mu_{N_{k}}\right)^3\right]}{\sigma^3_{N_{k}}}
% \]

This measure characterizes the asymmetry of the $k$-occurrence distribution $N_k$, and it is the metric most often used to estimate the presence of hubs. The more skewed the distribution of $N_k$, the greater the chance that a small number of vectors (hubs) will occur in the $k$-NN sets of other vectors. 

% Computing $S_{N_{k}}$ is empirically resource intensive for large datasets since it requires running exact search for each vector $\x \in \mathcal{D}$. 

\textbf{Synthetic and ANN Benchmark Datasets}: We use real and synthetic datasets as shown in Table~\ref{tab:hug-highway-exps-datasets} to study the Hub-Highway Hypothesis. In addition to a subset of ANN Benchmark datasets, we generate synthetic datasets by drawing vectors from the standard normal distribution. 

% To show empirical evidence supporting the Hub Highway Hypothesis, we start by examining the distribution of the number of times each node in the similarity search index is visited during search given a fixed number of queries, which we denote by $P_m(\x_i)$ to indicate that node $\x_i$ is visited exactly $m$ times during a fixed number of $k$-NN queries. We show that this distribution is skewed to the right, thus confirming that certain nodes ($\textit{highway nodes}$) are visited a disproportionate number of times. Next, using this distribution, we selectively choose the most visited nodes in the similarity search graph to be the hub node clusters and show evidence that these nodes are more connected to each other than random nodes using hypothesis tests. In the last experiment, we show that not only are these highway nodes more connected, they also allow for faster graph traversals for queries, hence supporting the claim that we no longer need hierarchy in high dimensional vector search where fast query times evolve as a result of the presence of the hub-highways. 

\subsection{Skewness of the Node Access Distribution}

The goal of this study is to support our claim that some nodes are visited far more frequently, and that this process is driven by hubs in the metric space. We examine the discrete distribution of the number of times each node in the index is visited during search given a fixed number of queries, which we write as $P_m(\x_i)$ (i.e., node $\x_i$ is visited $m$ times). If $P_m(\x_i)$  is right-skewed, it means that some nodes are very frequently visited.

\begin{table}[h]
\centering
\footnotesize
\caption{Similarity search index parameters}
\label{tab:indexparameters}
\begin{tabular}{cccc}   
\toprule
 $m$ & $ef$-construction & $ef$-search  & $k$\\
\midrule
32 & 100 & 200 & 100 \\
\bottomrule
\end{tabular}
\end{table}

Figure~\ref{fig:combined-kde-plots} shows the log-normalized node access count distribution for different datasets. We observe that as the dimension $d$ increases, this distribution becomes right-skewed for $\ell_2$ distance-based datasets. While this is strong evidence for the first part of our claim, it leaves open the possibility that some mechanism other than hubness is driving our observations. To control for this possibility, we also study the cosine distance, which is known to have anti-hub properties that prevent hub formation \cite{Radovanovi2010HubsIS}. We find that the cosine distance does not have a dramatic skew, even for $d \in \{1024, 1536\}$. The increased skewness of the $P_m(\x_i)$ distribution as $d$ increases demonstrates that highway nodes become increasingly prevalent as the dimension increases, and the differences between the $\ell_2$ and cosine results suggest that the hubness phenomenon is responsible for the formation of the highway nodes.

\subsection{Subgraph Connectivity of the Hub-Highway Nodes}
\label{section:subgraph-connectivity}

In this section we present empirical evidence confirming that hub-highway nodes exhibit strong connectivity in the graph. We begin by explaining our procedure to identify hubs. Let $\{\x_i\}_{i=1}^n$ be the vectors in a similarity search index for a dataset $\mathcal{D}$. 

\begin{itemize}
    \item We identify hub nodes as those that fall into the top percentile of the empirical node access distribution $P_m(\x_i)$. We use 95$^\text{th}$ and 99$^\text{th}$ percentile of node access counts as our threshold. We assign a binary label to each node to indicate whether it is a hub. Let $h: \mathcal{D} \to \{0, 1\}$ be this assignment function with $h(\x_i) = 1$ for nodes identified as hubs. 
    \item We wish to estimate the likelihood that a randomly-chosen out-neighbor of a hub node is, itself, a hub. To do so, we examine the 1-hop out-degree expansion of each hub and count the number of adjacent hub nodes. This yields a discrete distribution for the number of hubs to which each hub node is connected.
    \item Similarly, we select a set of random non-hub nodes from $V \coloneqq \mathcal{D} \setminus \bigcup_{i=1}^{n} h(\x_i) = 1$. For each node $\x_i \in V$, we compute the same quantity to find the number of hubs with which non-hubs are connected, allowing us to construct the equivalent distribution for non-hubs.
\end{itemize}

\vspace{-1.5em}

We test whether hubs differ from non-hubs in connectivity behavior by using statistical tests on the two distributions. Under the null hypothesis, both groups are equally likely to connect to other hubs; the alternative asserts that hubs preferentially attach to hubs.

We apply both a two-sample $t$-test and the Mann-Whitney U-test \cite{mann1947test}, the latter of which is more suitable for ANN datasets since it makes no normality assumption. With dataset sizes $n > 10^3$, Table~\ref{tab:pvalues_effect_sizes_p95} reports results using the top $5\%$ of nodes as hubs. At a 0.05 significance threshold, we reject the null in all but five cases using the U-test (and all but six with the $t$-test). Effect sizes are largest in synthetic $\ell_2$ datasets, likely due to their stronger hubness under this metric.

Using a stricter top-$1\%$ threshold (Table~\ref{tab:pvalues_effect_sizes_p99}), we reject the null in all but one case for both tests, with substantially larger effect sizes. These findings indicate that the most prominent hubs tend to form tightly connected subgraphs, consistent with our Hub Highway Hypothesis.

% Notably, the Yandex-DEEP benchmark does not exhibit the subgraph connectivity observed in most other benchmarks.
% Unlike the other embeddings in our collection, Yandex-DEEP was obtained by applying PCA and $\ell_2$ normalization to the last fully-connected layer of GoogleNet \cite{Szegedy2014GoingDW}. Theoretically, $\ell_2$ normalization should reduce the prevalence of hubs by restricting the distribution to the surface of the unit sphere. Yandex-DEEP also has the most uniform node access distribution of the $\ell_2$ datasets in Figure~\ref{fig:combined-kde-plots}. We believe that this may explain why Yandex-DEEP hubs are not as well-connected.
% % since the magnitude of the vectors does not disproportionately affect distance computations. 

\begin{table}[!ht]
\centering
\begin{minipage}[t]{0.49\textwidth}
    \centering
    \footnotesize
    \caption{Two-sample $t$-test and Mann-Whitney U-test results. Hub nodes are selected using the P95 threshold of the node access distribution.}
    \label{tab:pvalues_effect_sizes_p95}
    \adjustbox{width=\textwidth}{%
    \begin{tabular}{lcccc}
    \toprule
    Dataset & Dim & Mann-Whitney & Two-Sample $t$-Test & Effect Size \\
    \midrule
    IID Normal (Angular) & 16 & 0.3629 & 0.3090 & 0.0267 \\ 
    IID Normal (L2) & 16 & $\!<\!10^{-5}$ & $\!<\!10^{-5}$ & 0.3737 \\ 
    IID Normal (Angular) & 32 & 0.0335 & 0.0516 & 0.0872 \\ 
    IID Normal (L2) & 32 & $\!<\!10^{-5}$ & $\!<\!10^{-5}$ & 0.4275 \\ 
    IID Normal (Angular) & 64 & 0.0216 & 0.0148 & 0.1165 \\ 
    IID Normal (L2) & 64 & $\!<\!10^{-5}$ & $\!<\!10^{-5}$ & 0.3965 \\ 
    IID Normal (Angular) & 128 & 0.0083 & 0.0083 & 0.1284 \\ 
    IID Normal (L2) & 128 & $\!<\!10^{-5}$ & $\!<\!10^{-5}$ & 0.3773 \\ 
    IID Normal (Angular) & 256 & 0.0009 & 0.0007 & 0.1723 \\ 
    IID Normal (L2) & 256 & $\!<\!10^{-5}$ & $\!<\!10^{-5}$ & 0.2620 \\ 
    IID Normal (Angular) & 1024 & 0.1000 & 0.1114 & 0.0652 \\ 
    IID Normal (L2) & 1024 & $\!<\!10^{-5}$ & $\!<\!10^{-5}$ & 0.2361 \\ 
    IID Normal (Angular) & 1536 & 0.0957 & 0.1141 & 0.0645 \\ 
    IID Normal (L2) & 1536 & $\!<\!10^{-5}$ & $\!<\!10^{-5}$ & 0.2512 \\ 
    GloVe & 100 & $\!<\!10^{-5}$ & $\!<\!10^{-5}$ & 0.2550 \\ 
    NYTimes & 256 & $\!<\!10^{-5}$ & $\!<\!10^{-5}$ & 0.4488 \\ 
    GIST & 960 & $\!<\!10^{-5}$ & $\!<\!10^{-5}$ & 0.3645 \\ 
    Yandex-DEEP & 96 & 0.5002 & 0.5000 & 0.0000 \\ 
    Microsoft-SpaceV & 100 & 0.1586 & 0.1585 & 0.0535 \\ 
    \bottomrule
    \end{tabular}%
    }
\end{minipage}%
\hfill
\begin{minipage}[t]{0.49\textwidth}
    \centering
    \footnotesize
    \caption{Two-sample $t$-test and Mann-Whitney U-test results. Hub nodes are selected using the P99 threshold of the node access distribution.}
    \label{tab:pvalues_effect_sizes_p99}
    \adjustbox{width=\linewidth}{%
    \begin{tabular}{lcccc}
    \toprule
    Dataset & Dim & Mann-Whitney & Two-Sample $t$-Test & Effect Size \\
    \midrule
    IID Normal (Angular) & 16 & 0.0006 & 0.0006 & 0.1745 \\ 
    IID Normal (L2) & 16 & $\!<\!10^{-5}$ & $\!<\!10^{-5}$ & 0.6621 \\ 
    IID Normal (Angular) & 32 & 0.0347 & 0.0347 & 0.0972 \\ 
    IID Normal (L2) & 32 & $\!<\!10^{-5}$ & $\!<\!10^{-5}$ & 0.8173 \\ 
    IID Normal (Angular) & 64 & 0.0359 & 0.0417 & 0.0927 \\ 
    IID Normal (L2) & 64 & $\!<\!10^{-5}$ & $\!<\!10^{-5}$ & 0.8725 \\ 
    IID Normal (Angular) & 128 & 0.0093 & 0.0070 & 0.1316 \\ 
    IID Normal (L2) & 128 & $\!<\!10^{-5}$ & $\!<\!10^{-5}$ & 0.8428 \\ 
    IID Normal (Angular) & 256 & $\!<\!10^{-5}$ & $\!<\!10^{-5}$ & 0.3110 \\ 
    IID Normal (L2) & 256 & $\!<\!10^{-5}$ & $\!<\!10^{-5}$ & 0.8582 \\ 
    IID Normal (Angular) & 1024 & 0.1472 & 0.1318 & 0.0598 \\ 
    IID Normal (L2) & 1024 & $\!<\!10^{-5}$ & $\!<\!10^{-5}$ & 0.8314 \\ 
    IID Normal (Angular) & 1536 & $\!<\!10^{-5}$ & $\!<\!10^{-5}$ & 0.2356 \\ 
    IID Normal (L2) & 1536 & $\!<\!10^{-5}$ & $\!<\!10^{-5}$ & 0.8568 \\ 
    GloVe & 100 & $\!<\!10^{-5}$ & $\!<\!10^{-5}$ & 0.7642 \\ 
    NYTimes & 256 & $\!<\!10^{-5}$ & $\!<\!10^{-5}$ & 0.9305 \\ 
    GIST & 960 & $\!<\!10^{-5}$ & $\!<\!10^{-5}$ & 0.6829 \\ 
    Yandex-DEEP & 96 & 0.0013 & 0.0013 & 0.1614 \\ 
    Microsoft-SpaceV  & 100 & 0.0011 & 0.0011 & 0.1644 \\
    \bottomrule
    \end{tabular}%
    }
\end{minipage}
\end{table}

\subsection{Hub-Highway Nodes Enable Fast Traversal} 
% Allow Queries to Traverse the Graph Faster}

Our final question is whether the highway nodes allow queries to quickly traverse the similarity search graph.
While it is not surprising that a well-connected subgraph of frequently visited nodes would enable this behavior, it is not necessarily the case that queries would use the highway in the way predicted by our hypothesis, namely to quickly identify a neighborhood for deep exploration. To investigate this question, we track the sequence of nodes visited during beam search for several thousand queries. This allows us to determine the fraction of time spent within hub nodes in different phases of search.

Since beam search takes a variable number of steps for each query, we normalize by the total search length when presenting the results. More formally, suppose that $\x_1, \x_2, \hdots, \x_l$ is a length-$l$ sequence of such nodes visited by a query. We use the hub node assignment heuristic discussed in section \ref{section:subgraph-connectivity} to label $h(\x_i)$ each of these nodes as hubs / non-hubs. We then split the sequence into bins and compute the prevalence of hubs in each bin. For bin $B_i$, this is given by 
\[
\left(\frac{1}{|B_i|}\right)\sum_{\x_j \in B_i} h(\x_j)
\]
where $|B_i|$ is the bin size (fixed to $30$ in our analysis). By averaging this value over all queries, we can plot the likelihood of visiting a hub as the search progresses. 

\begin{figure}[htbp]
    \centering

    \begin{subfigure}[b]{0.48\linewidth}
        \centering
        \includegraphics[width=\linewidth]{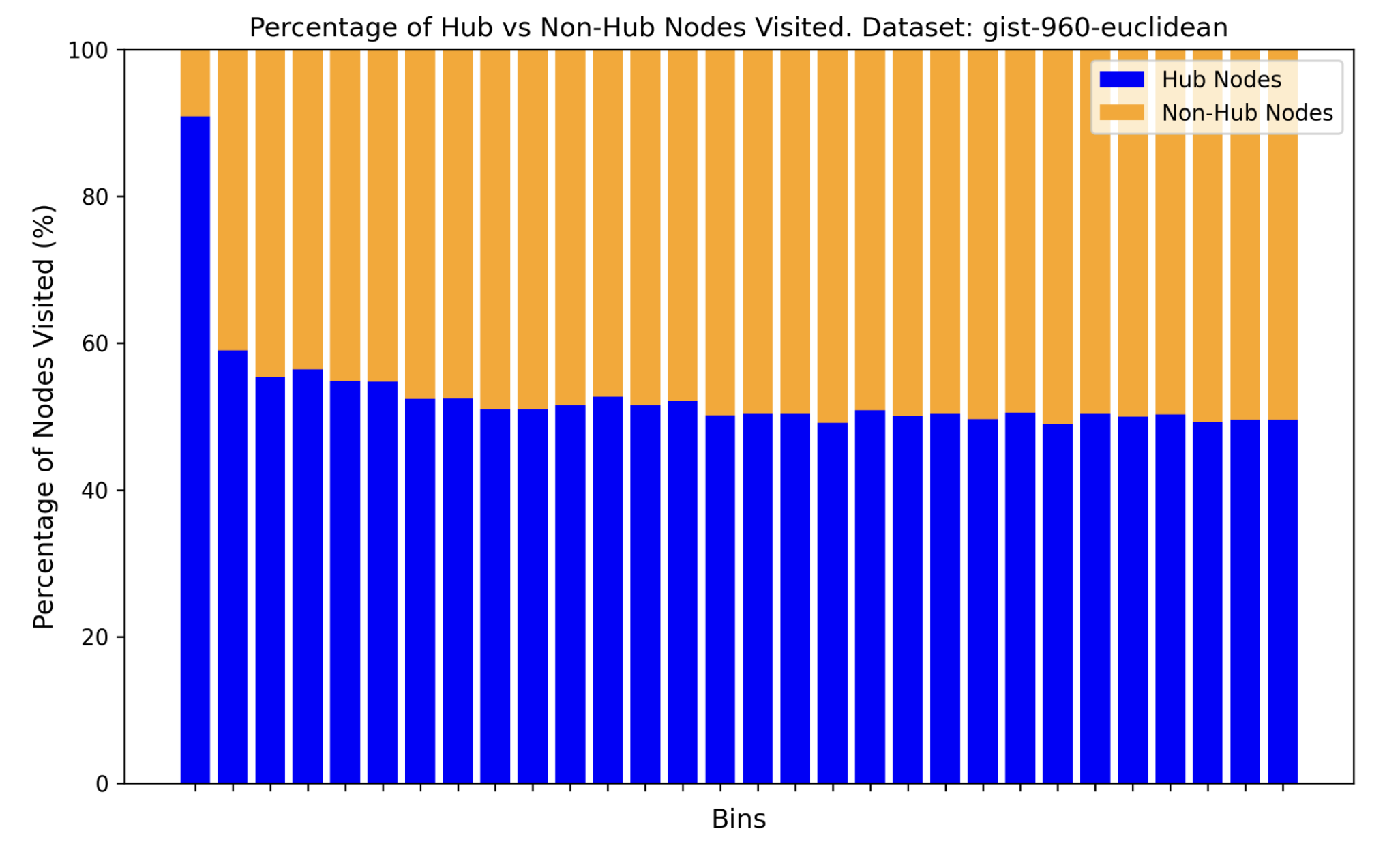}
        \caption{Gist}
        \label{fig:gist-speed-test}
    \end{subfigure}
    \hfill
    \begin{subfigure}[b]{0.48\linewidth}
        \centering
        \includegraphics[width=\linewidth]{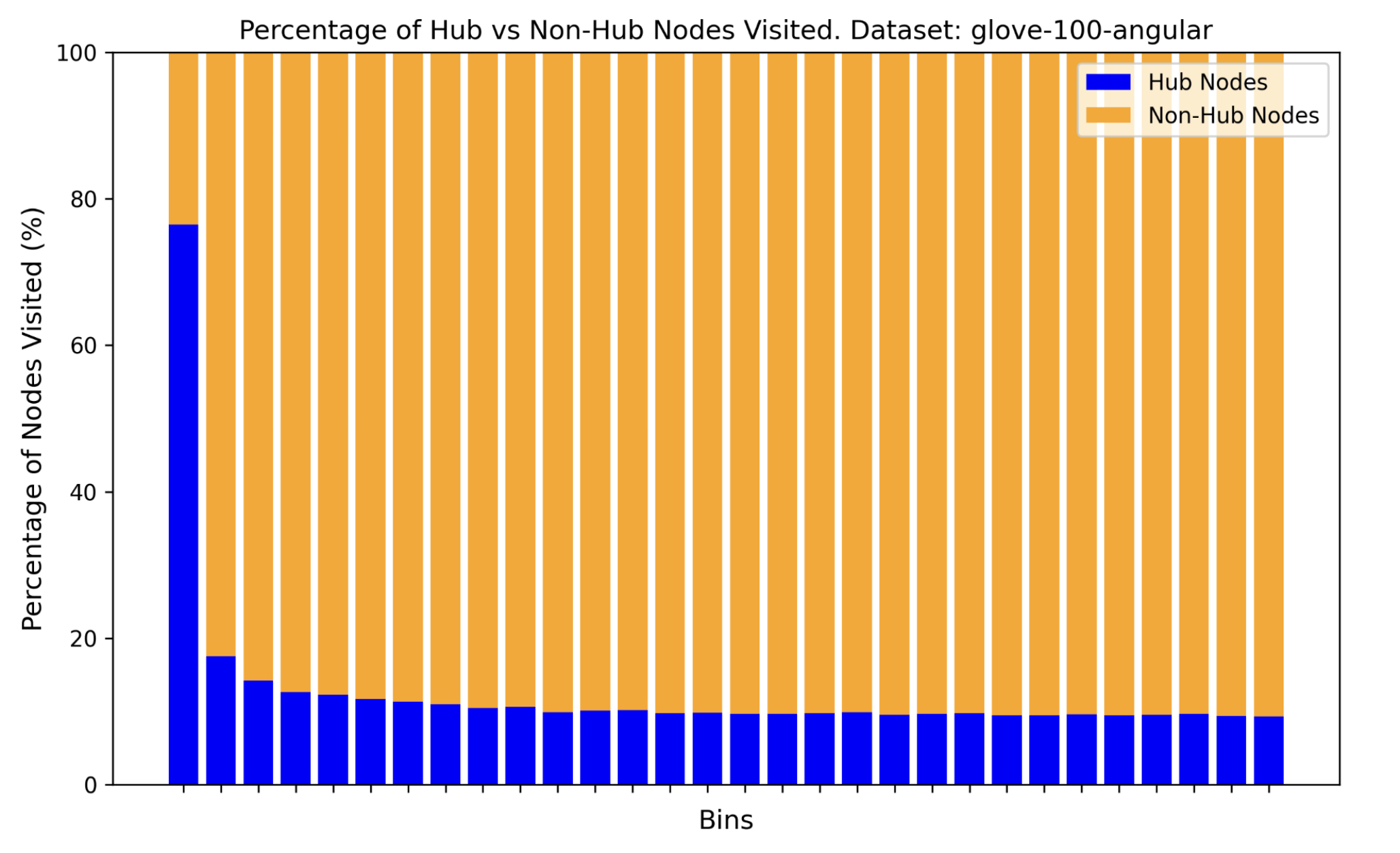}
        \caption{GloVe}
        \label{fig:glove-speed-test}
    \end{subfigure}

    \vspace{0.5em}

    \begin{subfigure}[b]{0.48\linewidth}
        \centering
        \includegraphics[width=\linewidth]{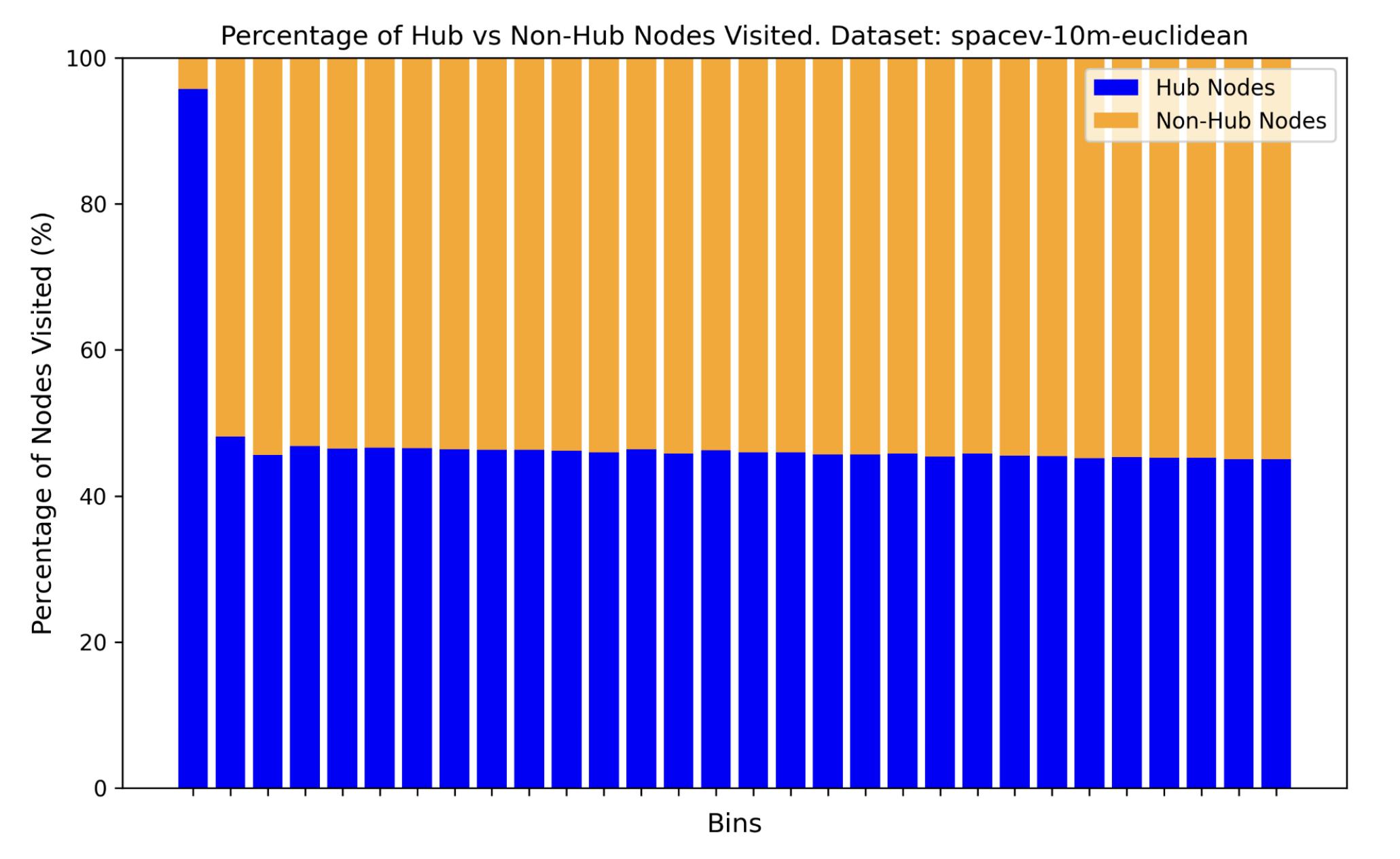}
        \caption{Microsoft SpaceV}
        \label{fig:spacev-speed-test}
    \end{subfigure}
    \hfill
    \begin{subfigure}[b]{0.48\linewidth}
        \centering
        \includegraphics[width=\linewidth]{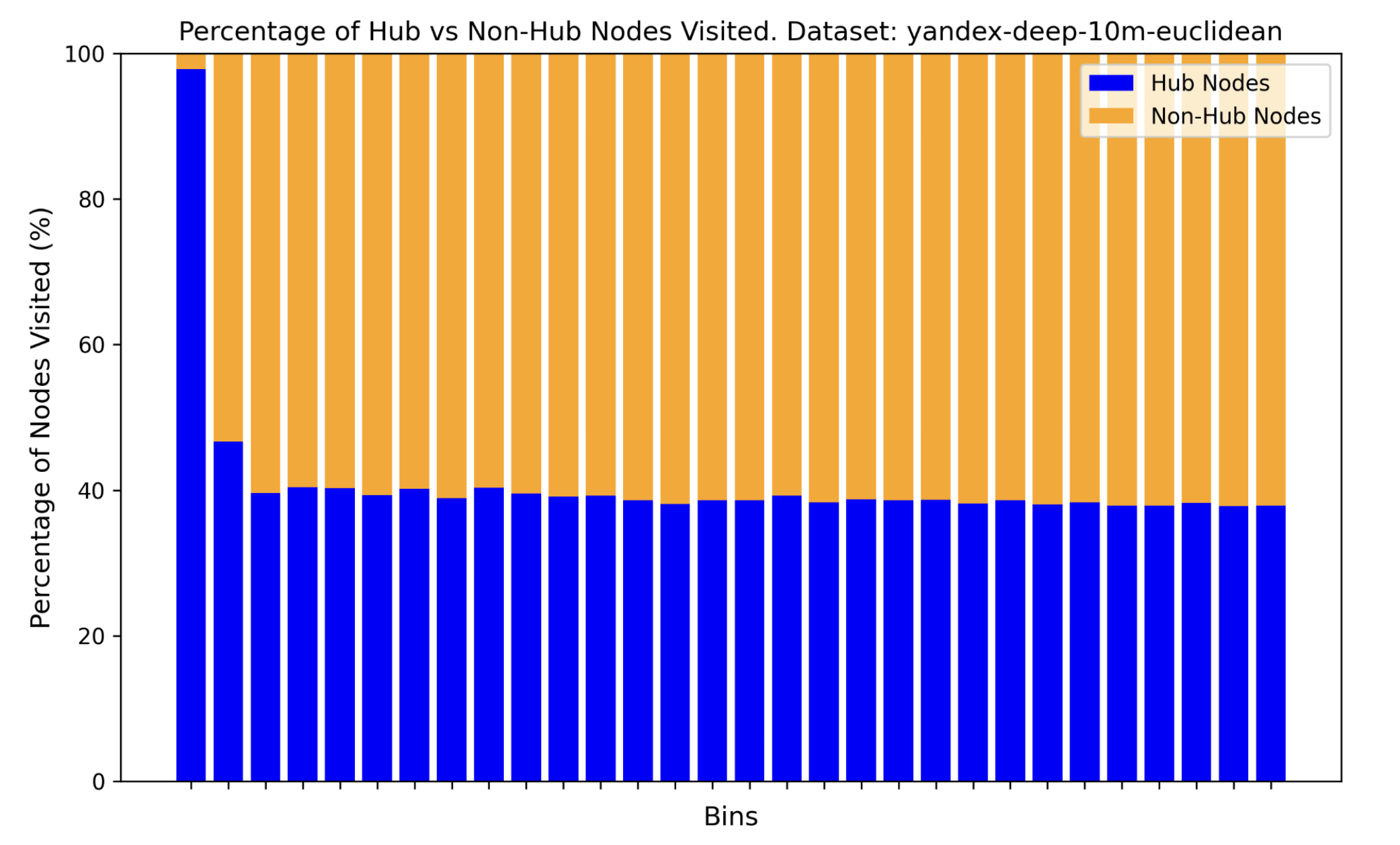}
        \caption{Yandex Deep}
        \label{fig:yandex-deep-speed-test}
    \end{subfigure}

    \caption{Highway nodes allow queries to traverse the graph faster.}
    \label{fig:combined-speed-test-plots}
\end{figure}

Figure~\ref{fig:combined-speed-test-plots} shows results for the Gist, GloVe, Microsoft SpaceV and Yandex-DEEP benchmark datasets. We observe that queries tend to concentrate in the highway structures early in search, shown by the high percentage of hub nodes visited in the first 5-10\% of the search steps. This result suggests that the highway allows queries to quickly navigate the similarity search graph until they find the region of the graph best suited for deep exploration. The propensity of the query to visit hubs appears to be tied to the hubness properties of the dataset. For example, the GloVe dataset uses the angular distance, has less pronounced hubs (Figure~\ref{fig:combined-kde-plots}), and queries spend a lower percentage of their time in hub nodes in this dataset (Figure~\ref{fig:glove-speed-test}). On the other hand, the GIST dataset has some of the highest highway utilization rates and is also our highest-dimensional $\ell_2$ dataset.

% This is consistent with prior findings that show that the cosine similarity exhibit anti-hub properties as it is not affected by the magnitude of the vectors compared with the $l_2$ distance.

% \begin{figure}[htbp]
%     \centering
%         \centering
%         \includegraphics[width=0.6\linewidth]{images/speed-tests/gist}
%         \caption{Gist}
%         \label{fig:gist-speed-test}
%         \centering
%         \includegraphics[width=0.6\linewidth]{images/speed-tests/glove}
%         \caption{GloVe}
%         \label{fig:glove-speed-test}
%         \centering
%         \includegraphics[width=0.6\linewidth]{images/speed-tests/spacev}
%         \caption{Microsoft SpaceV}
%         \label{fig:spacev-speed-test}
%         \centering
%         \includegraphics[width=0.6\linewidth]{images/speed-tests/yandex-deep}
%         \caption{Yandex Deep}
%         \label{fig:yandex-deep-speed-test}
%     \caption{Highway nodes allow queries to traverse the graph faster. }
%     \label{fig:combined-speed-test-plots}
% \end{figure}

\subsection{Discussion}

Our sequence of experiments provide substantial evidence supporting the \textit{Hub Highway Hypothesis}. Although it has long been established that the hubness phenomenon negatively affect common applications, such as clustering and even near neighbor search recall, existing ANNS methods, such as HNSW and NSG \cite{Zhao2023TowardsEI} mostly emphasize algorithmic improvements and performance optimizations only. We have shown that leveraging the inherent structures in the data, particularly hub-highway occurrences, should be central to the design of new scalable similarity search indexes. 

\textbf{Scientific Implications:}
Our work reveals that the hub highway is an intrinsic and naturally-forming structure in high-dimensional proximity graphs. We believe this observation is both novel and has important implications for the scaling potential of many traversal heuristics.
Specifically, we expect the benefits of sophisticated search initialization methods to decay under hub-style preferential attachment.

Initialization is a recurring and popular research direction for graph-based near-neighbor search, and leading algorithms vary greatly in their initialization techniques.
The research question dates back to the seminal 1993 paper by Arya and Mount~\cite{arya1993approximate}, which conjectured that clever search initialization -- in their case, via $kd$-tree -- could improve performance over random initialization.
Over the following three decades, the research community has investigated diverse stratified sampling based on clusters~\cite{sebastian2002metric}, vantage-point trees~\cite{iwasaki2010proximity}, seeds formed from the graph expansion of $kd$-trees~\cite{iwasaki2018optimization} and previously-visited nodes~\cite{wang2012query}, hierarchical graphs~\cite{Malkov2016EfficientAR}, hierarchical clustering~\cite{munoz2019hierarchical}, dataset medoids~\cite{Subramanya2019DiskANNF}, and several other methods before finally returning to cluster-stratified candidates~\cite{oguri2024theoretical, ni2023diskannplusplus} and random entry nodes~\cite{jaiswal2022ood}.

How is it that these early papers on graph search show performance gains, even as today's best vector databases return to simple initializations without performance loss? Our hub highway hypothesis offers a clear explanation for this apparent contradiction: In the early 2000s and 2010s, datasets were low-dimensional and initialization was important to avoid local minima and long graph detours. However, modern vector databases contain data that is sufficiently high-dimensional to naturally form a fast-routing structure, explaining why initialization no longer drives performance. Based on our results, we conjecture that the largest algorithmic improvements to graph-based ANNS should come from optimizations that affect the connectivity and cost of traversal in the base graph, such as link pruning and search algorithm design.

\section{Conclusion}

Approximate near neighbor search has become an increasingly
crucial computational workload in recent years with the seminal
Hierarchical Navigable Small World (HNSW) algorithm continuing
to garner significant interest and adoption from practitioners. We present the first comprehensive study on the utility of the hierarchical component of HNSW over numerous large-scale datasets and performance metrics. Ultimately, we find that the hierarchy of HNSW provides no clear benefit on high-dimensional datasets and can be \textbf{removed} without any discernible loss in performance while providing memory savings. 

While similar observations have been made before in the literature \cite{lin2019graph, coleman2022graph, dobson2023scaling}, we are, to our knowledge, the first to conduct an exhaustive study over modern benchmark datasets and taking extensive care to compare implementations with performance engineering parity. Furthermore, we go beyond prior works and study \emph{why} the hierarchy does not help, culminating in our introduction of the \emph{Hub Highway Hypothesis}, an empirical result on how proximity graphs built over high-dimensional metric spaces leverage a small subset of well-connected nodes to traverse the network quickly. We believe that our results provide immediate implications for practitioners seeking to save memory or simplify their vector database implementations and we look forward to further partnering with the community on these endeavors. 

% One limitation of our work, however, is that we do not provide any theoretical formalism to explain this why this highway structure accelerates search. More generally, there is still a sizable gap between theory and practice in graph-based near neighbor search, and we believe further exploration along these lines could be a valuable direction for future work.  

\subsection*{Acknowledgments}
This material is based upon work supported by the U.S. National Science Foundation under Grant No. 2313998. Any opinions, findings, and conclusions or recommendations expressed in this material are those of the author(s) and do not necessarily reflect the views of the U.S. National Science Foundation.

\bibliographystyle{icml2025}
\bibliography{icml2025/example_paper}
\clearpage

\appendix

\section*{Appendix}

\section{Background: Similarity Search \& HNSW}
\subsection{Similarity Search}

In the similarity search (or $k$-NNS) problem, we are interested in retrieving $k$ elements from a dataset $\mathcal{D} = \{\x_i,\hdots, \x_n\} \subset \mathbb{R}^d$ that minimize the distance to a given query $q \in \mathbb{R}^d$ (or, equivalently, maximize the vector similarity). More precisely, given a similarity function $\phi : \mathbb{R}^d \times \mathbb{R}^d \to \mathbb{R},$ the nearest neighbor $\x^* \in \mathcal{X}$ of $q$ is defined as
\[
\x^* \coloneqq \argmax_{\x_{i} \in \mathcal{D}} \phi(\x_i,q)
\]

where $\phi$ is usually the $\ell_2$ or cosine similarity. With the enormity of modern data workloads and the underlying vector dimensionality, it becomes computationally infeasible to exhaustively search for the true top-$k$ neighbors for any query $q.$ Thus, approximate search algorithms trade-off quality of the search for lower latency. 

In the approximate nearest neighbor search (ANNS) regime, we evaluate the quality of the search procedure typically by the Recall$@k$ metric. More formally, suppose a given ANNS search algorithm outputs a subset $\mathcal{O} \subseteq \mathcal{D}, |\mathcal{O}| = k, $ and let $G \subseteq \mathcal{D}$ be the true $k$ nearest neighbors of a query $q.$ We define Recall$@k$ by $\frac{|\mathcal{O} \cap G|}{k}$. ANNS algorithms seek to maximize this metric while retrieving results as quickly as possible. 

\subsection{HNSW Overview}
\label{section:hnsw-overview}

With this formalization of the ANNS problem, we will now briefly review the key elements of the HNSW algorithm, which is the central focus of our benchmarking study. As we alluded to previously, HNSW builds off of prior work in \emph{navigable small world graph} indexes introduced in \cite{malkov2014approximate}. Small world graphs are a well-studied phenomenon in both computing and the social sciences and are primarily defined by the fact that the average length of a shortest path between two vertices is small (typically scaling logarithmically with the number of nodes in the network) \cite{travers1977experimental, watts1998collective, kleinberg2000navigation}. Small world graphs are also often characterized by the presence of well-connected \emph{hub nodes} which we discuss further in the next section. 

While small world graphs are, by construction, suited for efficient greedy graph traversal, the HNSW authors argue that the polylogarithmic scaling of the search process is still too inefficient for the demands of near neighbor search on large datasets. This claim motivates the design of HNSW where the hierarchy allows for computing a fixed number of distances in each graph layer independent of the network size.

% HNSW Construction
Specifically, the HNSW index is constructed in an iterative fashion. For a newly inserted element $x$, the algorithm will randomly select a maximum layer $l$ and then insert the new point into every layer up to $l$. This randomized process is executed with an exponentially decaying probability distribution such that, in expectation, each subsequent layer has exponentially more nodes than its predecessor. Within a layer, HNSW greedily adds edges between $x$ and its $M$ closest neighbors (where $M$ is a hyperparameter) where the neighbors consist of previously inserted points. This process then repeats in the subsequent layer below using the closest neighbors found in the prior graph as entry points. Through this process, the top layer of the hierarchy will be the coarsest directed graph, consisting of the fewest nodes and edges, and the bottom layer will be the densest and contain all of the nodes, each with connections to (up to) $M$ neighbors. As an additional, and important, optimization, HNSW also implements the pruning heuristic of \cite{arya1993approximate} that will prune an edge from $u$ to $v$ if there exists another edge from $u$ to a neighbor $w$ of $v$ such that the distance from $u$ to $w$ is less than that of $u$ to $v$. 
%HNSW Search

The search procedure of HNSW, described in Algorithm \ref{alg:query} also executes iteratively where the algorithm maintains a list of candidate points at each layer of the hierarchical graph before returning the final list $k$ nearest neighbor candidates after traversing the base graph layer.

\begin{algorithm}
\caption{HNSW Construction}
\begin{algorithmic}[1]

\State \textbf{Input:} Set of data points $D$, max layer $L_{max}$, max connections per layer $M$, layer insertion probability $m_l$, size of dynamic candidate list $efc$
\State \textbf{Output:} HNSW graph with hierarchical layers

\Procedure{Construct}{$D, L_{max}, M, m_l, efc$}
    \State Initialize empty hierarchical graph $G$
    \State Initialize entry point $ep \gets \text{None}$
    \For{each $p \in D$}
        \State $L_p \gets$ GeometricDistribution($m_l$)
        \If{$ep = \text{None}$}
            \State Set $p$ as entry point $ep$
            \State Insert $p$ into all levels $\leq L_p$
        \EndIf
        \For{$l = L_{max}$ to $L_p$}
            \State $ep \gets$ SearchLayer($G, l, p, ep, efc$) \Comment{Algorithm \ref{alg:query}}
        \EndFor
        
        \For{$l = 0$ to $L_p$}
            \State $N \gets$ SelectNeighbors($p, G, l, M$)
            \State Add edges from $q$ to each neighbor $n \in N$ at layer $l$
            \If{$n\in N$ has $< M$ edges}
            \State Add back-connections to $q$ to node $n$.
            \Else
            \State Run SelectNeighbors on $\{q, $ edges of $n\}$.
            \EndIf
            \EndFor
        \EndFor
        
        \If{$L_p > L_{ep}$}
            \State $ep \gets p$
        \EndIf
\EndProcedure

\State

\Function{SelectNeighbors}{$p, G, l, M$}
    \State Compute distances from $p$ to all nodes in $G[l]$
    \State Return $M$ nodes based on selection heuristic in ~\cite{arya1993approximate}
\EndFunction

\end{algorithmic}
\end{algorithm}

%\subsection{HNSW Search Algorithm}

\begin{algorithm}
\caption{HNSW Query}
\label{alg:query}
\begin{algorithmic}[1]
\State \textbf{Input:} Graph $G$, layer $l$, query $q$, starting point $p$, number of nearest neighbors to return $efs$
\Procedure{SearchLayer}{$G, l, q, p, efs$}
\State Candidate queue $C = {p}$, currently top
results queue $T = {p}$, visited list $V = {p}$

\While{$C$ is not empty}

\State{$c \gets \text{nearest element from}~C~\text{to}~q$}

\State{$f \gets \text{furthest element from}~T~\text{to}~q$}

\If{dist(c, q) > dist(f, q)}
\Return T
\EndIf

\For {$e \in \text{neighbourhood}(c)~\text{at layer}~l$}

\If{$e \in V$}
\State continue
\EndIf

\State{$V.add(e)$}
\If{$dist(e, q) \le dist(f, q)~\text{or}~|T| \le efs$}
\State{$C.add(e)$}
\State{$T.add(e)$}
\EndIf

\If{$|T| \ge efs$}

\State{Remove furthest point to q from T}
\EndIf

\State{$f \gets \text{furthest element from}~T~\text{to}~q$}
\EndFor

\Return T

\EndWhile
\EndProcedure
\end{algorithmic}
\end{algorithm}

\section{Reproduction of Prior Studies}
\label{section:repro-of-prior-studies}

In this section, we present a replicability study using four \texttt{flatnav} NSW implementation. In particular, we revisit the experimental design of two prior works in the literature: the original 2016 HNSW paper of \cite{Malkov2016EfficientAR} and a subsequent 2019 paper from \cite{lin2019graph} that found limitations with the hierarchical component of HNSW. As we discussed in the previous section, these prior works possess limitations in experimental design, scope of benchmarking datasets, and a lack of analysis into understanding the results, which motivates our work in this paper. Nevertheless, we use these prior studies as a starting point to see if we can independently replicate these results via our own FlatNav implementation. Such a reproduction would both further validate the soundness of these previous experiments over the test of time as well as provide confirmation of the correctness of FlatNav before we proceed to new, larger-scale benchmarks.

Following the same setups as \cite{Malkov2016EfficientAR} and \cite{lin2019graph} we generate a series of random vector datasets of varying dimensionality where each vector component is sampled uniformly at random from the range $[0, 1)$. In particular, we consider dimensionalities of $d=4, 8, 16$ and 32. As in \cite{lin2019graph}, we set the number of near neighbors to retrieve to $k=1$ (departing from the default of $k=100$ we use elsewhere in this paper). We also tried including the \texttt{sw-graph} NSW baseline \cite{boytsov2013engineering} that \cite{Malkov2016EfficientAR} used in their evaluation to benchmark against HNSW, but we were not able to run this older library successfully. However, we were able to replicate these prior findings using our own \texttt{flatnav} implementation which is conceptually identical to \texttt{sw-graph} but with more software optimizations to achieve engineering parity with \texttt{hnswlib}.

\begin{figure}[htbp]
    \centering
    \includegraphics[width=\linewidth]{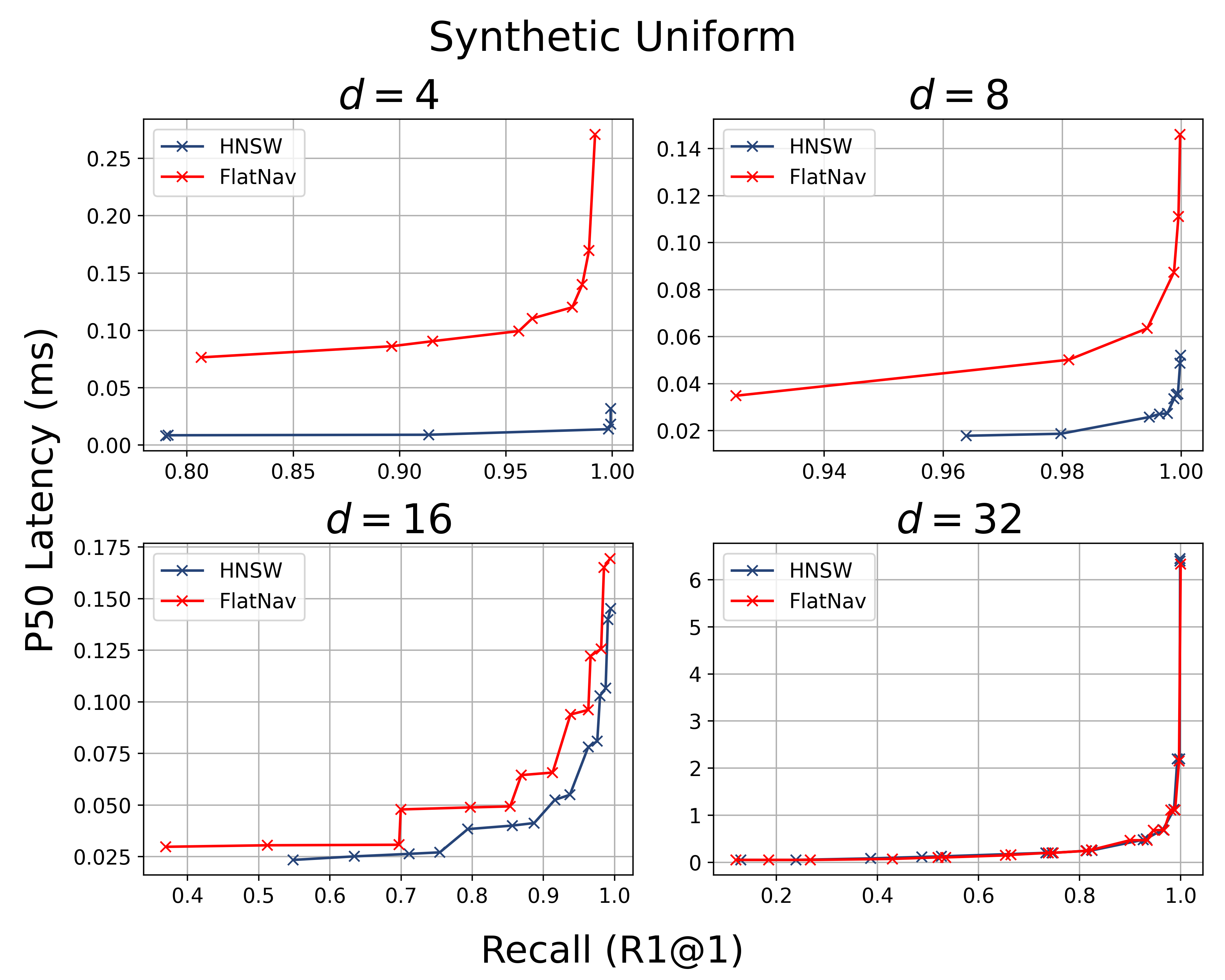}
    
    \caption{Median Latency vs. Recall of HNSW and FlatNav across dimensions $d=4, 8, 16, 32$. We observe that the hierarchical structure accelerates search only when $d < 32$, matching the findings of \cite{lin2019graph}. Our results demonstrating a significant advantage with HNSW on synthetic datasets with dimensionality $d=4$ and $d=8$ also match the findings of the original HNSW paper \cite{Malkov2016EfficientAR}}.
    \label{fig:linzhao}
\end{figure}

As shown in Figure \ref{fig:linzhao}, we successfully replicated the experiments benchmarking HNSW versus a flat NSW graph from two prior research papers. Notably, both of these previous works primarily experiment with randomly generated vector data with very low dimensionality by the standards of modern machine learning. Coupled with our findings in the next section where we find no discernible difference between HSNW and a flat graph index on high-dimensional datasets, our results suggest a simple decision criterion for selecting a search index: \textbf{For dimensionality $d < 32$, HNSW and the hierarchy provide a speedup. Otherwise, the simplicity and memory savings of a flat NSW index provide more benefit.}

We also had the opportunity to discuss our findings with the lead author of \cite{Malkov2016EfficientAR} who confirmed that the hierarchy provides a robust speedup on these low-dimensional datasets but noted the performance on higher dimensional vectors remained less clear, which further motivated us to take up the benchmarking study in the next section.

\section{Dataset Statistics}
\subsection{Benchmark Datasets for the Latency-Recall Tradeoff}

Table~\ref{tab:datasets-exps} shows the different benchmark datasets used in Section~\ref{sec:main-datasets}. 

\begin{table}[ht]
\centering
\scriptsize
\begin{tabular}{llll}
\toprule
\textbf{Dataset} & \textbf{Dimensionality} & \textbf{\# Points} & \textbf{\# Queries} \\
\midrule
BigANN$^\dagger$  & 128 & 100M & 10K \\
Microsoft SpaceV$^\dagger$  & 100 & 100M & 29.3K \\
Yandex DEEP$^\dagger$  & 96  &  100M  & 10K \\
Yandex Text-to-Image$^\dagger$ & 200 &  100M & 100K \\
GloVe  & \{25, 50, 100, 200\}  & 1.2M & 10K \\
NYTimes  & 256   & 290K  & 10K\\
GIST  & 960  & 1M  & 1K \\
SIFT  & 128 & 1M  & 10K \\
MNIST  & 784  & 60K & 10K \\
DEEP1B & 96   & 10M  & 10K \\
\bottomrule
\end{tabular}
\caption{Dataset Statistics. The datasets marked by $\dagger$ are from the BigANN benchmarks \cite{Simhadri2022}. The remaining are taken from ANN Benchmarks \cite{Aumller2018ANNBenchmarksAB}.}
\label{tab:datasets-exps}
\end{table}

\subsection{Benchmark Datasets for the Hub-Highway Hypothesis Experiments}

Table~\ref{tab:hug-highway-exps-datasets} details the various ANN and Big-ANN benchmark datasets as well as the synthetic datasets used in Section~\ref{section:hubs} for illustrating the empirical evidence of the hub-highway hypothesis. 

\begin{table}[h]
\centering
\scriptsize
\caption{Hub-Highway Experimental Datasets}
\label{tab:datasets} 
\begin{tabular}{lccc}
\toprule
Dataset & Dimensionality & \# Points & \# Queries \\
\midrule
GIST & 960 & 1M & 1k \\
GloVe & 100 & 1.2M & 10k \\
NYTimes & 256 & 290K & 10k \\
Yandex-DEEP & 96 & 10M & 10k\\
Microsoft-SpaceV & 100 & 10M & 29.3k \\
IID Normal & \{$2^4$, $2^5$, $2^6$, $2^7$, $2^8$, $2^{10}$, $1.5 \cdot 2^{10}$\} & 1M & 10k \\
IID Normal & \{$2^4$, $2^5$, $2^6$, $2^7$, $2^8$, $2^{10}$, $1.5 \cdot 2^{10}$\} & 1M & 10k \\
\bottomrule
\end{tabular}
\label{tab:hug-highway-exps-datasets}
\end{table}

\section{Extended Discussion and Limitations}

\subsection{Extended Discussion}
\textbf{Small-World Graphs:} The network science research community has known for decades that long-range connections and hubs induce the formation of ``small-world'' graphs that are easily traversed~\cite{watts1998collective}. 
This idea has been enormously influential in ANNS, providing the motivation for both NSW and HNSW~\cite{malkov2012scalable, malkov2014approximate}, but our results suggest that ANN graphs constructed over low-dimensional datasets may not in fact exhibit small-world properties.
Because the $k$-occurrence distribution is near-uniform for intrinsically low-dimensional data distributions, a pure $k$NN graph (without pruning or long-range links) will not create hubs with a high in-degree.
We believe that hierarchical structures are helpful in low dimensions because they help to induce hub behavior, by ensuring that search always begins from a small set of nodes.
However, this is not necessary to produce hubs and induce small-world properties in high dimensions. The $k$NN graph construction process is sufficient on its own, because the hub highway emerges in high dimensions.

\subsection{Limitations}

Despite our notable empirical evidence for the hug-highway hypothesis and its significance in understanding graph-based ANN search, a principled understanding of this phenomenon from theoretical grounds is still lacking. Future research efforts can be directed towards explicit bounds on the probability of a query $q$ reaching high latency on a proximity graph $G = (V,E)$ consisting of hub nodes $H$. 

There are several reasons why such a theoretical formulation is hard to attain. First, query latency is a metric that is the most perturbed by both engineering optimizations, such as SIMD operations, as well as the underlying hardware. Therefore, even if one attains a theoretical framework for bounding the expected query latency on a proximity graph, it is still very plausible that pure engineering optimizations could realize better performance in practice. Second, even if we control for engineering optimizations and the underlying hardware, it is not sufficient to consider the most obvious factors including graph complexity (i.e., $|V|$ and $|E|$) and the dimensionality of the underlying vector space $d$. Different graph construction procedures and pruning algorithms will induce different expected latency bounds. Future research directed to this theoretical understanding of the hub-highway hypothesis will prove to be invaluable to both theoreticians and practitioners alike. 

\section{Extended Benchmarks}
\label{extended-bench}
\subsection{Building FlatNav from Scratch}

In the main body of the paper, we present a series of results demonstrating that there is essentially no difference in performance between search over HNSW with a full hierarchy and search over the flat NSW base graph. As we discuss in Section \ref{latency-benchmarks}, we fix the experimental design in the main body of the paper to extract the base NSW graph from the \emph{same} \texttt{hnswlib} code that constructs the full HSNW index and use our \texttt{flatnav} implementation of the search algorithm to traverse the graph. In other words, we construct the full hierarchical graph with \texttt{hnswlib} and then extract the base layer as our separate flat graph index. We designed the benchmarking experiments in this manner to avoid any potential confounding effects from differences in code between the baseline HNSW graph construction and our own version. 

However, this experimental setup raises a separate concern over whether the hierarchical component of HNSW might still be useful to \emph{construct} the base graph index even if the hierarchy is not used during search. In this extended benchmark, we provide additional results to demonstrate that this is not the case. In particular, when we benchmark the full hierarchical HNSW index from \texttt{hnswlib} against a flat graph built from scratch with no hierarchy at all via our \texttt{flatnav} library, we observe identical results to what we report in Section \ref{latency-benchmarks}. Thus, we can conclude that the hierarchical component of HNSW does not seem to be useful for high-dimensional workloads for either construction or search. 

\begin{figure}[htbp]
  \centering

  % Row 1: Two latency plots
    \centering
    \includegraphics[width=0.5\textwidth]{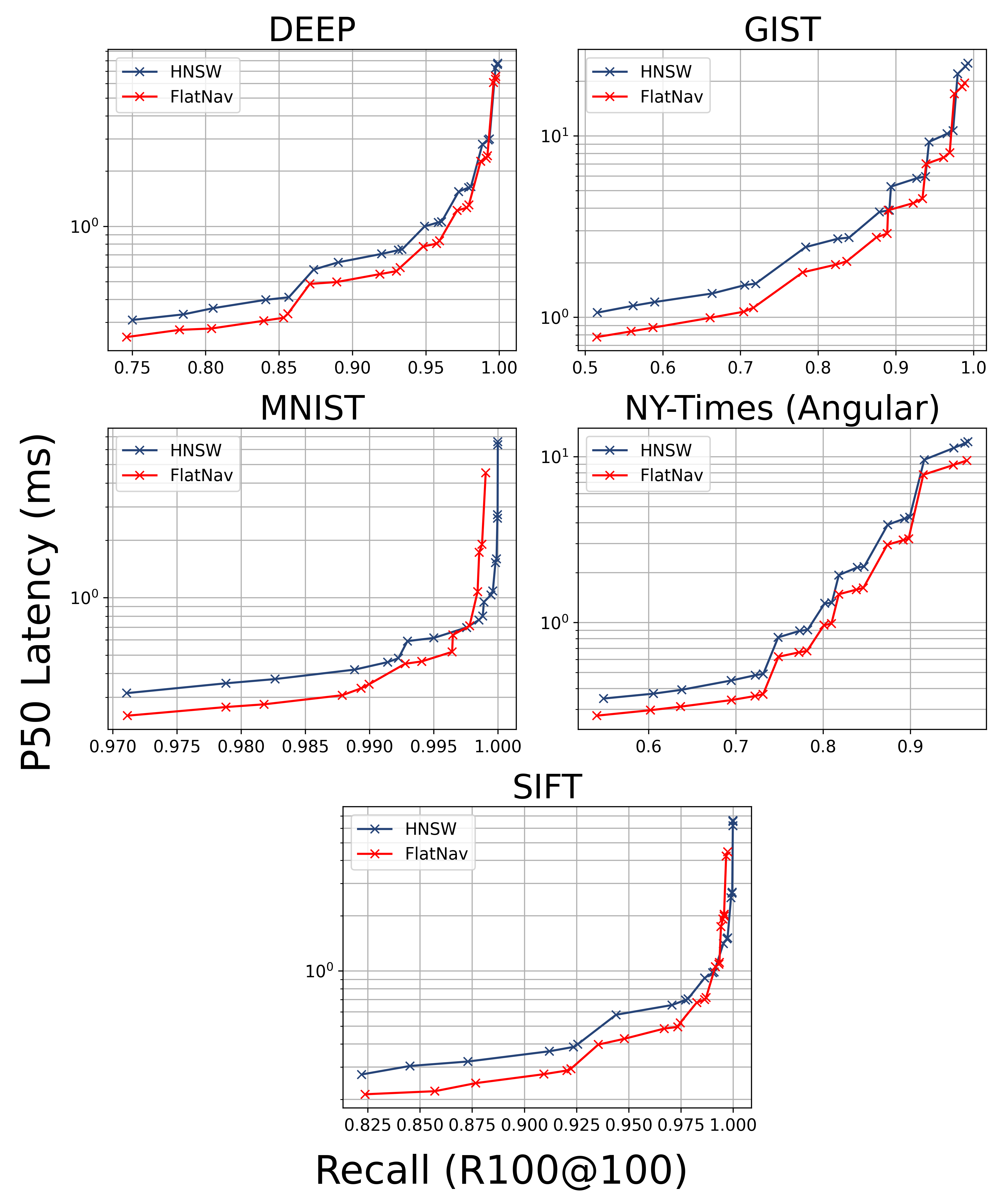}
    \caption{The p50 Latency vs Recall relationship between HNSW and FlatNav (constructed from scratch) is identical to the relationship between HNSW and FlatNav  (base layer extracted from HNSW) shown in Figure \ref{fig:annbench-p50}.}
    \label{fig:annbench-p50-appendix}
  \hfill
    \centering
    \includegraphics[width=0.5\textwidth]{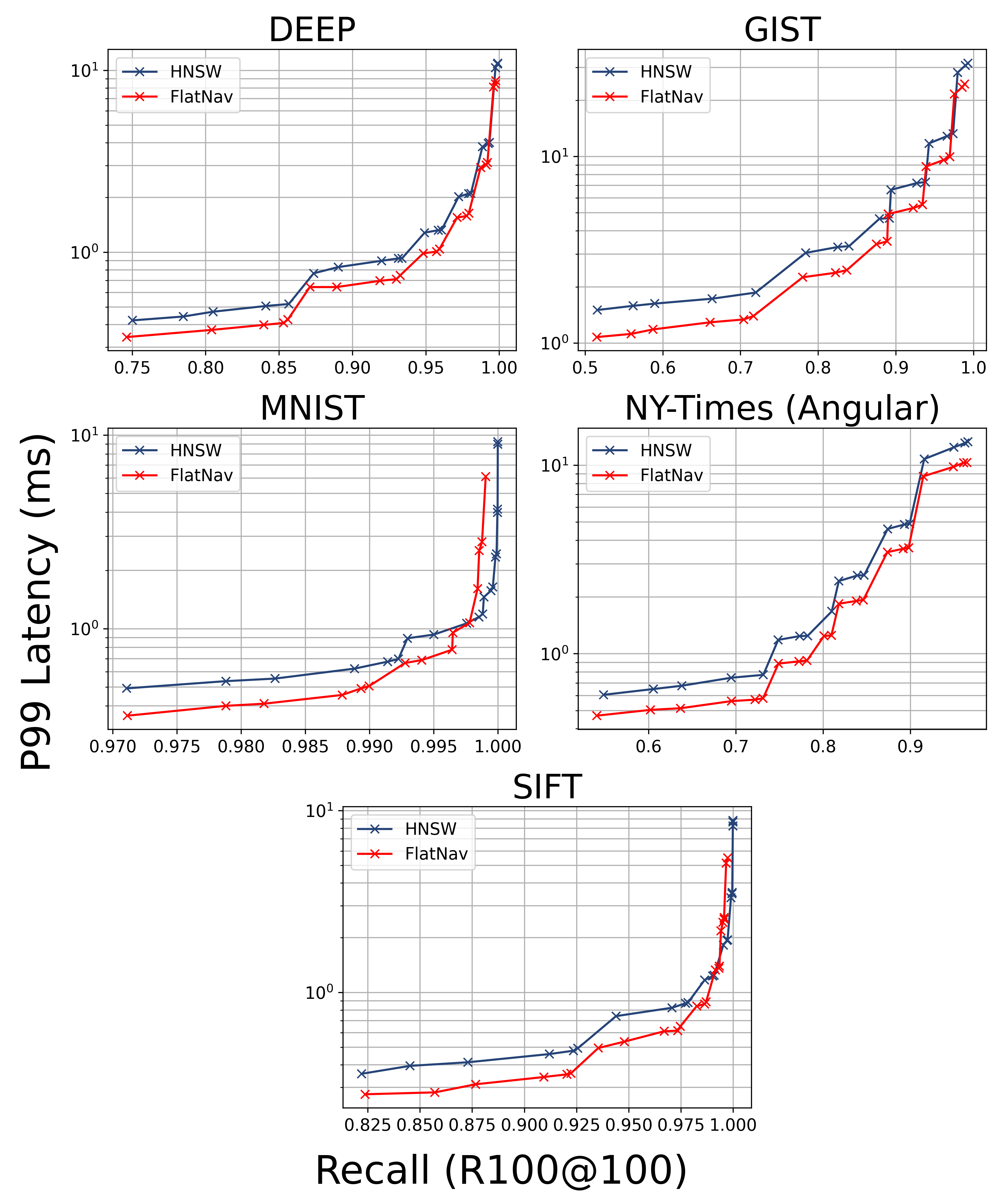}
    \caption{The p50 Latency vs Recall relationship between HNSW and FlatNav (constructed from scratch) is identical to the relationship between HNSW and FlatNav (base layer extracted from HNSW) shown in Figure \ref{fig:annbench-p99}.}
    \label{fig:annbench-p99-appendix}

  \vspace{1em} % Small vertical gap between rows

\end{figure}

In Figures \ref{fig:annbench-p50-appendix} and \ref{fig:annbench-p99-appendix} we show the benchmarking results of HNSW and FlatNav with the latter built from scratch with no hierarchy even in construction. The results are identical to our findings in Figures \ref{fig:annbench-p50} and \ref{fig:annbench-p99} in Section \ref{latency-benchmarks} in the main body of the paper. For these extended benchmarks, we use a cloud server equipped with an AMD EPYC 9J14 96-Core Processor and 1 TB of RAM.

\section{Extended Hubness Experiments}

\subsection{Extending the Hub-Highway Hypothesis to LLM Embeddings }

\begin{figure}[t]
  \centering
  \includegraphics[width=0.8\linewidth]{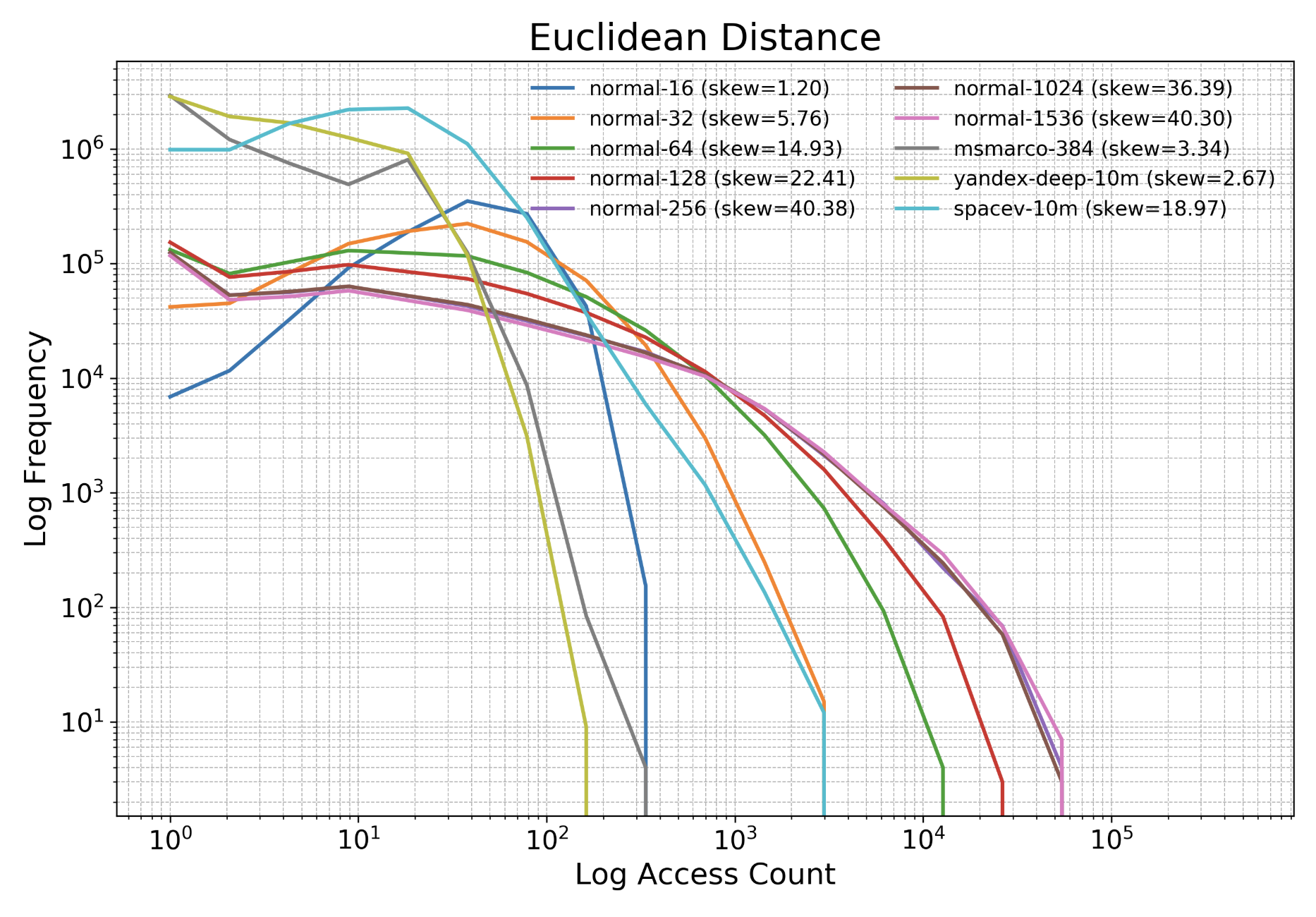}
  \caption{Log-normalized node access count distribution $P_m(\x_i)$ for $\ell_2$ datasets along with MSMARCO embeddings.}
  \label{fig:msmarco-log-log-plot}
\end{figure}

To evaluate whether the Hub-Highway Hypothesis generalizes beyond sythetic gaussian-distributed and ANN benchmark datasets, we apply the same analysis to large language model (LLM) embeddings used in information retrieval. In particular, we examine the MSMARCO \cite{Campos2016MSMA} dataset, a widely-used retrieval benchmark comprising millions of real-world queries and documents. 

\textbf{Data generation.} We encode all training split queries in 
MSMARCO using the all-MiniLM-L6-v2 model from the \texttt{SentenceTransformers} \cite{Reimers2019SentenceBERTSE}
library. This yields 384-dimensional vector representations. 
Using these embeddings, we construct a $k$-NN proximity graph (using $k = 100$) 
and compute the node access distribution $P_m(\x_i)$, defined as the 
number of times node $\x_i$ is visited across a fixed set of queries 
using HNSW beam search heuristic. As with all experiments, we fix the number of queries to be 10,000.

\textbf{Results.} Figure~\ref{fig:msmarco-log-log-plot} shows the 
log-normalized node access frequency distribution for MSMARCO. 
The distribution exhibits a long-tail behavior similar to what we observed in synthetic $\ell_2$ datasets and the Big-ANN benchmarks, with a clear skew indicating that a small subset of nodes are accessed orders of magnitude more frequently than others. This supports our hypothesis that highway-like structures emerge naturally even in real-world retrieval embeddings generated by LLMs.

These findings suggest that the routing behavior of hub nodes is not an artifact of synthetic data or benchmark construction, but a general phenomenon in high-dimensional embedding spaces. This reinforces our claim that such hubs can effectively replace the hierarchy in modern ANN graph indexes.

\subsection{No Hierarchy Memory Savings}

\begin{table}[ht]
\centering
\footnotesize
\begin{tabular}{llll}
\toprule
\textbf{Dataset} & \textbf{Dataset Size} & \textbf{\texttt{hnswlib}} & \textbf{\texttt{flatnav}} \\
\midrule
BigANN  & 100M & 183 & 113 \\
Microsoft SpaceV  & 100M& 104 & 85.5  \\
Yandex DEEP  & 100M& 100  &  60.7 \\
\bottomrule
\end{tabular}
\caption{Peak Index Construction Memory in GBs. We observe that \texttt{flatnav} requires considerably less memory during construction compared to \texttt{hnswlib}.}
\label{tab:datasets-memory}
\end{table}

Section~\ref{latency-benchmarks} focused on demonstrating that we can remove the hierarchy in HNSW with impunity for latency benchmarks. Here we turn our attention to memory consumption and measure the memory savings from removing the hierarchy by running a memory profiler during index construction. In terms of memory allocation, similarly to \texttt{flatnav}, \texttt{hnswlib} allocates static memory during index construction comprising base layer node allocation, a visited node list and a list of mutexes in the multi-threaded setting. Additionally, it also incurs memory cost attributable to the hierarchy, particularly maintaning the dynamically allocated links between nodes at each layer.

We benchmarked both libraries against a subset of the BigANN benchmarks. Table~\ref{tab:datasets-memory} shows the peak memory allocated by the two implementations during index construction for the BigANN, Microsoft SpaceV and Yandex DEEP benchmarks. Since multithreading has a runtime overhead, we fix the number of cores to 32 in each one of the stated benchmark. For BigANN, we observe a $38\%$ reduction in peak memory, a $39\%$ reduction for Yandex DEEP, and an $18\%$ reduction for the Microsoft SpaceV benchmark. This shows that we are able to save significant memory by removing the hierarchy, and it is likely that we can optimize \texttt{flatnav} implementation to save memory further. 

One caveat of these reported memory savings is that we are comparing different two software implementations in \texttt{hnswlib} and \texttt{flatnav}. Since \texttt{hnswlib} is a mature library widely used by practitioners as well as researchers, it supports more features than \texttt{flatnav} and thus must maintain additional complexity whereas our implementation, while performant, is more of a research prototype. Therefore, differences in code may account for a significant part of the peak memory usage differences. Nevertheless, we believe our findings are relevant and still noteworthy given that \texttt{hnswlib} is so widely adopted. By demonstrating that we can considerably reduce the memory overhead of \texttt{hnswlib} without sacrificing performance, we hope to bring the community's attention to the opportunities for further optimization in this direction. 

\subsection{Hub formation: Discerning Metric Space Hubness from Preferential Attachment }
While Figure~\ref{fig:combined-kde-plots} confirms the existence of hub nodes in the dataset, it does not distinguish between hubs that arise from the properties of the underlying metric space and hubs that form through some other mechanism. It is possible that \textit{preferential attachment} explains the formation of hubs, since NSW graphs are built incrementally by sequentially adding points to an existing graph. Nodes that are added early in graph construction may become hubs by accumulating a greater-than-average share of inbound graph links, rather than by being popular neighbors in the metric space.

To investigate the effects of preferential attachment, we computed the variance ($R^{2}$ of a linear model) in the empirical node access distribution explained by the insertion ordering(Table~\ref{tab:preferential_attachment}). We log-transformed both the node access count and the insertion order before running the linear model and confirmed that the residuals are approximately normal after examining the QQ plots ($p < 10^{-6}$ for all models).

We observe a modest effect from the node insertion order in our synthetic data. This is particularly true for the angular datasets, which we believe to be due to the weaker hubness phenomena produced by the angular distance metric. Preferential attachment may account for a relatively greater share of the node access distribution when metric hubs are not present to heavily skew the distribution.

Ideally, we would repeat this analysis using the $K$-occurrence distribution, to show that the hubness of the metric space is more strongly predictive of the node acccess count than the insertion order. Unfortunately, it is not feasible to compute the $K$-occurrence distribution due to the $O(n^2)$ brute-force computation cost. However, we believe that the results in Table~\ref{tab:preferential_attachment} still support the idea that the dimensionality of the metric space strongly contributes to the formation of hubs in the \textit{Hub-Highway Hypothesis}, especially when combined with the evidence in Figure~\ref{fig:combined-kde-plots}.

% \begin{table}[h!]
\begin{table}[!htbp]
\centering
\footnotesize
\caption{Correlation analysis for node insertion order and node access count, to determine effects of preferential attachment.}
\label{tab:preferential_attachment}
\adjustbox{width=\linewidth}{%
\begin{tabular}{lcccc}
\toprule
Dataset & Dimension & Explained Variance (\%) \\
\midrule
IID Normal (Angular) & 16 & $3.0$\% \\
IID Normal (Angular) & 32 & $8.7$\% \\
IID Normal (Angular) & 64 & $16.6$\% \\
IID Normal (Angular) & 128 & $23.3$\% \\
IID Normal (Angular) & 256 & $24.6$\% \\
IID Normal (Angular) & 1024 & $24.1$\% \\
IID Normal (Angular) & 1536 & $23.9$\% \\
IID Normal (L2) & 16 & $< 0.1$\% \\
IID Normal (L2) & 32 & $3.1$\% \\
IID Normal (L2) & 64 & $7.6$\% \\
IID Normal (L2) & 128 & $8.7$\% \\
IID Normal (L2) & 256 & $7.1$\% \\
IID Normal (L2) & 1024 & $7.1$\% \\
IID Normal (L2) & 1536 & $6.6$\% \\
\midrule
GloVe (Angular) & 100 & $0.2$\% \\
NYTimes (Angular) & 256 & $0.3$\% \\
GIST (L2) & 960 & $< 0.1$\% \\
Yandex-DEEP (L2) & 96 & $0.3$\% \\
Microsoft-SpaceV (L2) & 100 & $0.5$\% \\
\bottomrule
\end{tabular}%
}
\end{table}

\end{document}